\pdfoutput=1
\documentclass[11pt,onecolumn]{cleantechnicalreport}
\usepackage[authoryear,sort&compress,round]{natbib}
\usepackage{hybridfrontmatter}
\usepackage{graphicx}

\makeatletter

\newcommand{\ReportHuggingIcon}{%
\raisebox{-0.2em}{%
\includegraphics[height=1em]{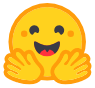}%
}%
}

\newcommand{\report@hugging}{}
\newcommand{\reporthugging}[1]{%
\gdef\report@hugging{#1}%
\global\report@hasmetadatatrue
}

\patchcmd{\abscontent}
{\ifdefempty{\report@contact}}
{%
\ifdefempty{\report@hugging}{}{%
\report@metarow{\ReportHuggingIcon}{Hugging Face}{%
\href{\report@hugging}{\nolinkurl{\report@hugging}}%
}%
}%
\ifdefempty{\report@contact}%
}
{}
{%
\PackageError{reporthugging}
{Could not insert the Hugging Face link}
{Check the definition of abscontent in hybridfrontmatter.sty.}%
}

\makeatother

\usepackage{listings,textcomp}
\usepackage{tcolorbox}
\usepackage{multirow}
\tcbuselibrary{listings,skins}

\lstdefinestyle{svuprompttext}{
language={},
basicstyle=\linespread{1}\fontsize{8.5}{9.5}\selectfont\rmfamily,
columns=fullflexible,
keepspaces=true,
breaklines=true,
breakatwhitespace=true,
showstringspaces=false,
upquote=true,
tabsize=2,
aboveskip=0pt,
belowskip=0pt
}

\newtcblisting[auto counter]{svuprompt}[2][]{%
enhanced,
listing only,
listing engine=listings,
title={Prompt~\thetcbcounter: #2},
colback=black!4,
colbacktitle=black!16,
colframe=black!25,
coltitle=black,
fonttitle=\small\bfseries\linespread{1}\selectfont,
boxrule=0.5pt,
arc=1.5mm,
outer arc=1.5mm,
boxsep=0.5mm,
left=2mm,
right=2mm,
top=1mm,
bottom=1mm,
toptitle=0.8mm,
bottomtitle=0.8mm,
before skip=6pt,
after skip=6pt,
listing options={style=svuprompttext},
#1
}

\makeatletter
\renewcommand{\report@leftlogo}{%
\includegraphics[height=15mm]{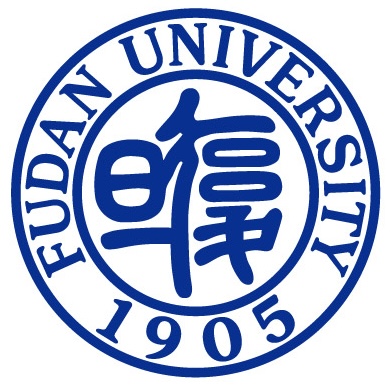}%
\hspace{2mm}%
\includegraphics[height=15mm]{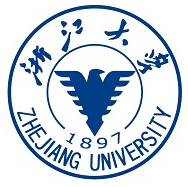}%
\hspace{2mm}%
\includegraphics[height=15mm]{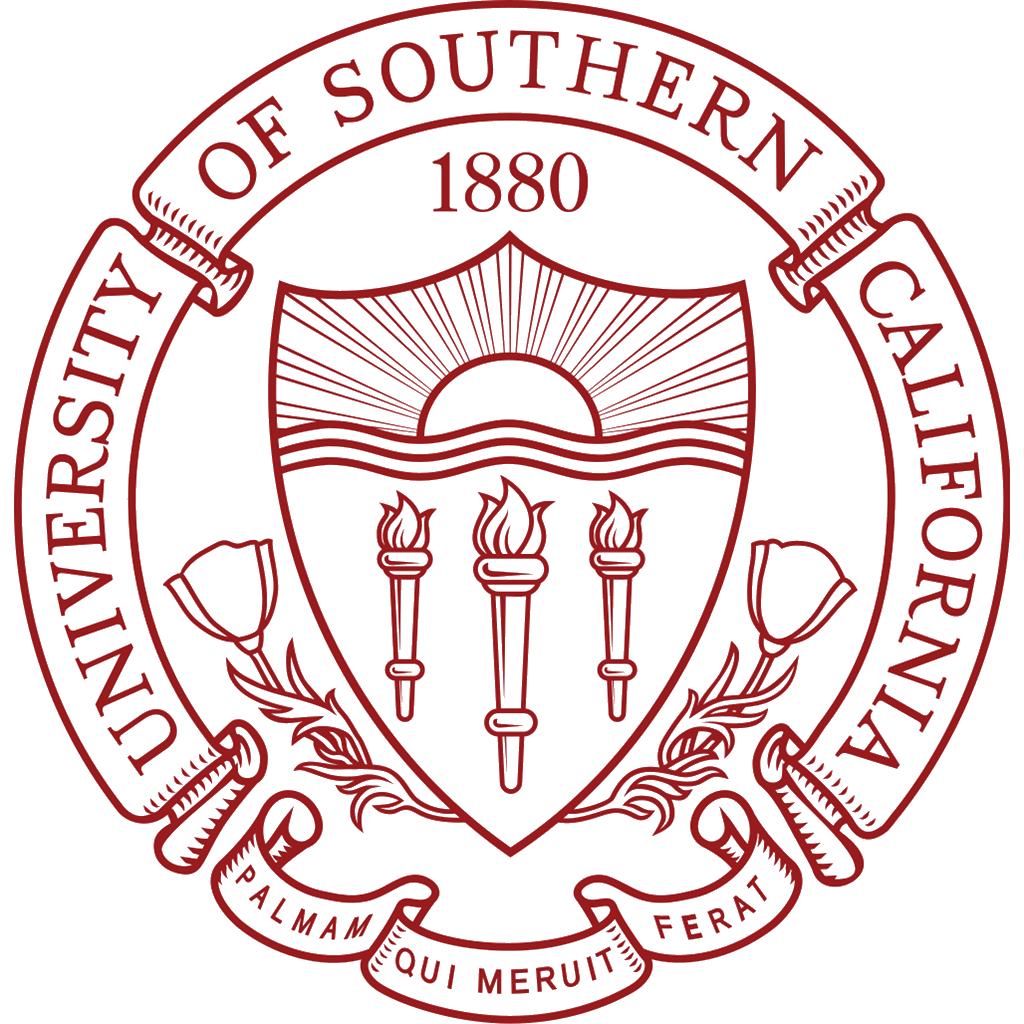}%
\hspace{2mm}%
\includegraphics[height=15mm]{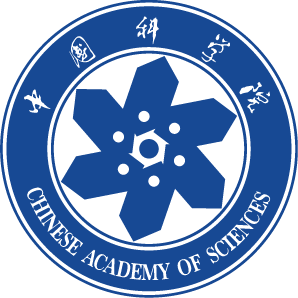}%
\hspace{2mm}%
\includegraphics[height=15mm]{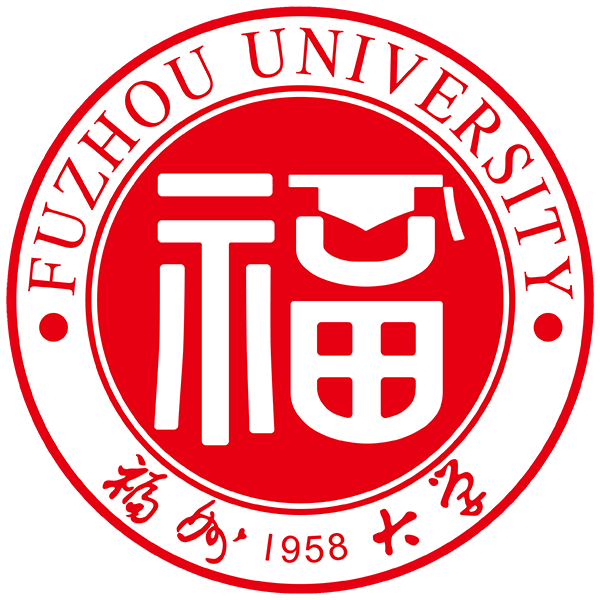}%
\hspace{2mm}%
\includegraphics[height=15mm]{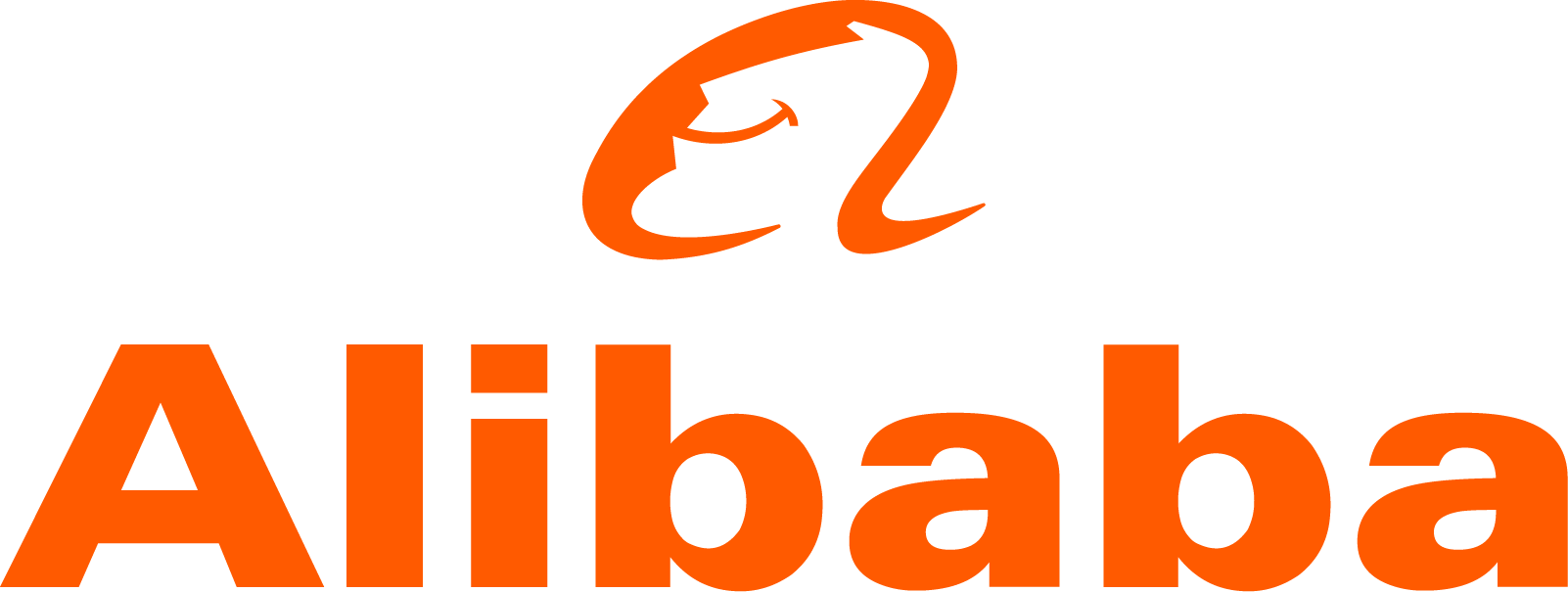}%

}

\makeatother

\title{%
\texorpdfstring{%
\mbox{%
\raisebox{-0.18em}{%
\includegraphics[height=1.25em]{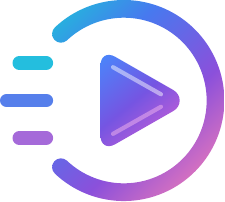}%
}%
\hspace{0.25em}SVUBench%
}%
}{SVUBench}:
A Comprehensive Benchmark for Proactive Streaming Video Understanding%
}

\author[1]{Yuxuan Liang*}
\author[1]{Bohan Zhang*}
\author[1]{Haiyang Yu*\ddag}
\author[2]{Bocheng Hu}
\author[1]{Ke Niu}
\author[1]{Xu Li}
\author[3]{Zhengtao Yao}
\author[3]{Yan Wen}
\author[1]{Haotian Chen}
\author[1]{Xiaolei Chen}
\author[4]{Jiacong Wang}
\author[1]{Zhe Liu}
\author[]{Jingqun Tang}
\author[5]{Benlei Cui}
\author[6]{Fan Shi$^{\dagger}$}
\author[1]{Xiangyang Xue}

\affil[1]{Fudan University}
\affil[2]{Zhejiang University}
\affil[3]{University of Southern California}
\affil[4]{University of the Chinese Academy of Sciences}
\affil[5]{Alibaba Group}
\affil[6]{Fuzhou University}

\reportauthornote{$^{*}$ Equal Contribution}

\reportauthornote{$^{*}$ Equal Contribution \quad $^{\ddagger}$ Project Lead \quad $^{\dagger}$ Corresponding author}

\reporthugging{https://huggingface.co/datasets/yuxuanliang/SVUBench}
\reportwebsite{https://lyxxx1211.github.io/SVUBench}

\begin{abstract}
Proactive streaming video understanding requires models to interpret unfolding events, respond timely, and retrieve relevant historical evidence. To systematically evaluate these capabilities, we introduce SVUBench, a comprehensive proactive streaming video benchmark comprising \textbf{12,131} questions and \textbf{62,881} answer points across ten proactive tasks spanning Perception, Understanding, and Reasoning. Constructed through model-assisted annotation and human verification, \textbf{SVUBench introduces two complementary interaction modes}. Basic evaluates sustained proactive responses from video onset, while Trajectory requires historical recovery and continuous response updates following task activation during streaming. A \textbf{Duration-Scaled Challenge} further evaluates these capabilities across four video duration tiers. We develop the \textbf{Proactive Response Score (PRS)} with task-adaptive answer windows to jointly assess answer correctness, response timeliness, historical consistency, and content quality. Extensive experiments and analysis reveal the strengths and limitations of existing models, highlighting directions for advancing proactive streaming video understanding. 
\end{abstract}

\begin{document}

\maketitle

\section{Introduction}
\label{sec:intro}

% \begin{figure*}[t]
%   \centering
%   \includegraphics[width=1\textwidth]{visualization/intro_v3.pdf}
%   \caption{\textbf{Overview of SVUBench.} Left: hierarchical taxonomy of ten proactive tasks across three cognitive dimensions. Right: three key components of the benchmark, including (i) hierarchical task scope and the Duration-Scaled Challenge, (ii) proactive response evaluation with task-adaptive response windows and metrics, and (iii) dual interaction modes, Basic and Trajectory, for evaluating sustained responses, historical recovery, and continuous updates.}
%   \label{fig:intro}
% \end{figure*}

The rapid development of large language models (LLMs) \citep{bai2023qwen,chowdhery2023palm,touvron2023llama,guo2025deepseek-r1} has driven the emergence of multimodal large language models (MLLMs), extending their capabilities to visual perception, reasoning, and video understanding \citep{qwenvl,liu2024llava,blip2,llava1.5}. However, most Video MLLMs operate offline, processing complete videos before responding \citep{video-llava,llava-video,chen2025longvila,llavaov}. Growing demands for continuous perception and interaction have therefore driven a shift toward streaming video understanding, where models process visual input incrementally \citep{VideoLLM-online,Flash-VStream}. Beyond reactive responses to user queries, recent work further explores proactive streaming models that autonomously determine when to respond as events unfold, enabling more timely assistance \citep{zhang2026progressiveonline,yao2025timechat,azad2026streamready}.

Despite these advances, reliable proactive streaming understanding remains challenging. First, diverse tasks, from state change detection and event counting to relational reasoning, require broad capabilities in perception, understanding, and reasoning \citep{lin2026streamingbench,niu2025ovobench}. Second, models must determine when sufficient evidence supports a response, avoiding premature or delayed answers while continuously monitoring new events and updating their responses \citep{VideoLLM-online,azad2026streamready}. Finally, growing observation histories demand effective memory management to preserve and retrieve relevant past events within their temporal context \citep{jiang2026memento,zhang2026hermes,song2024moviechat,qian2024videostreaming,chatterjee2025providellm,chen2026streamkv}. These challenges highlight the need for a comprehensive benchmark that systematically evaluates task competence, response timeliness and continuity, and historical retrieval in proactive streaming scenarios.

\begin{figure}[t]
  \centering
  \includegraphics[width=1\linewidth]{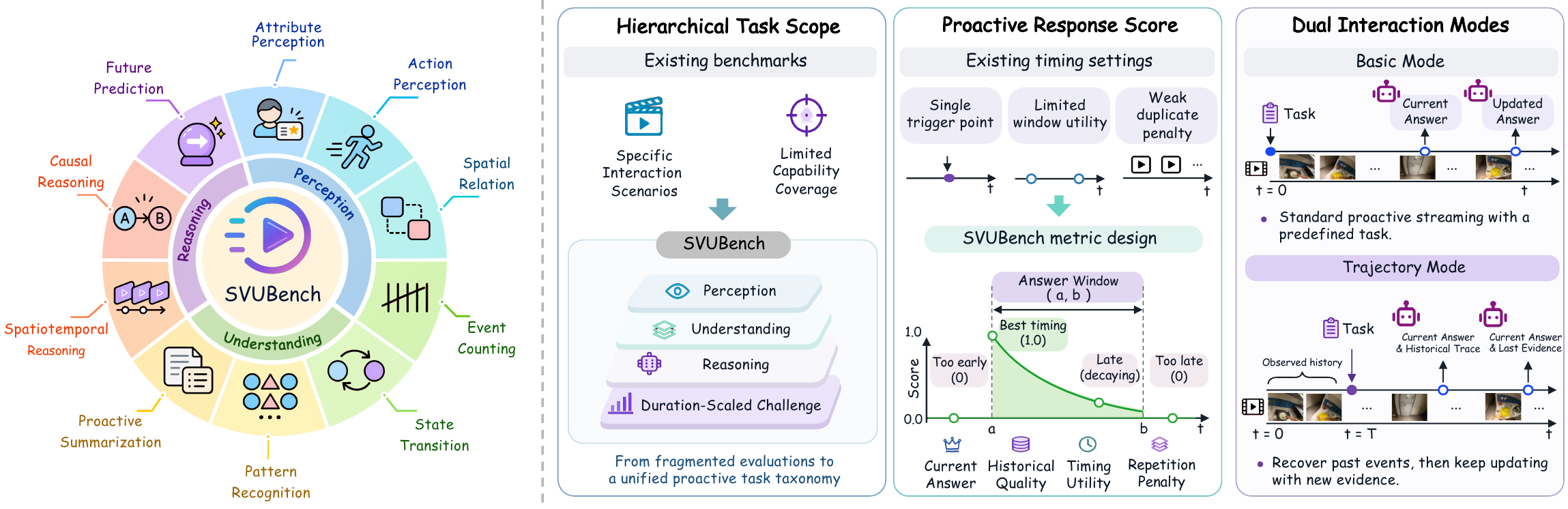}
  \caption{\textbf
{Overview of SVUBench.}  Left: hierarchical taxonomy of ten proactive tasks across three cognitive dimensions. Right: (a) hierarchical task scope and the Duration-Scaled Challenge, (b) proactive response score with task-adaptive answer windows, and (c) Basic and Trajectory modes for evaluating sustained responses, historical recovery, and continuous updates.}
  \label{fig:intro}
\end{figure}

Existing streaming video benchmarks have substantially advanced the evaluation of temporal understanding and proactive interaction \citep{lin2026streamingbench,niu2025ovobench,wang2025proactivevideoqa,zhao2026omnipro}. However, their evaluation remains limited in task coverage and sustained proactive responses across multiple events \citep{liu2026svcbench,shi2026river}. In particular, jointly evaluating proactive segment captioning and final summarization can provide a more comprehensive assessment of continuous perception, historical memory, and information integration. Furthermore, existing timing protocols often rely on fixed tolerances around annotated timestamps, without adequately accounting for differences in evidence requirements and valid response windows across tasks \citep{wang2025proactivevideoqa,xun2026rtvbench,azad2026streamready}. Finally, retrospective question answering alone cannot fully assess whether models can recover relevant historical trajectories and sustain proactive responses as new evidence emerges \citep{huang2025videochatonline,lin2026streamingbench,shi2026river}. These limitations highlight the need for a unified benchmark that systematically evaluates diverse cognitive capabilities, continuous proactive responses, historical memory management, and task adaptive response timing.

To address these limitations, we introduce SVUBench, a comprehensive benchmark for proactive streaming video understanding, as illustrated in Figure \ref{fig:intro}. It comprises 3,398 samples, 12,131 question points, and 62,881 answer points across ten fine-grained tasks spanning three cognitive dimensions, namely \textbf{Perception, Understanding, and Reasoning}. This taxonomy enables systematic evaluation of models' abilities to perceive ongoing events, track evolving contexts, and reason over accumulated evidence. We further incorporate proactive captioning and final summarization to assess the understanding of successive video segments and the integration of salient information into a coherent global summary. In addition, \textbf{Duration-Scaled Challenge} organizes long videos into four duration tiers to examine model capabilities over increasingly extended histories.

SVUBench further adopts two complementary interaction modes, \textbf{Basic and Trajectory}, to evaluate proactive understanding under different task activation conditions. Basic activates tasks at video onset and assesses sustained responses to unfolding events, while Trajectory introduces tasks during streaming and requires models to recover relevant historical events and continuously update their answers as new evidence emerges. We develop task-adaptive response windows and the\textbf{ Proactive Response Score (PRS)} to assess response quality and timeliness, complemented by joint evaluation of current answers and historical trajectories and semantic and temporal assessment of captioning and summarization. Together, these components provide a comprehensive evaluation of response accuracy, timeliness, historical consistency, and content quality. Extensive experiments on representative models further reveal their capabilities and limitations, offering insights into future directions for proactive streaming video understanding. Our main contributions are summarized as follows.

\begin{itemize}
    \item \textbf{Hierarchical Task Taxonomy.}
    We introduce ten proactive tasks across three cognitive dimensions and a four-tier Duration-Scaled Challenge.

    \item \textbf{Dual Interaction Modes.}
    We propose Basic and Trajectory modes to evaluate sustained responses, historical recovery, and continuous updates.

    \item \textbf{Proactive Response Evaluation.}
    We develop Proactive Respons Score with task-specific answer windows to evaluate quality, timeliness, and historical consistency.

    \item \textbf{Dataset Construction and Model Evaluation.}
    We construct 3,398 samples with 12,131 questions and 62,881 answer points through model-assisted annotation and human verification, and systematically evaluate representative models to reveal their capabilities and limitations, providing insights for future research.
\end{itemize}

\section{Related Work}
\label{sec:related}

\textbf{Video Understanding Models.}
Offline video understanding models have advanced video captioning and question answering \citep{bai2025qwen3,bai2025qwen2.5,wang2025internvl3.5,video-llava,llava-video,llavaov}. Growing demands for continuous perception have driven the development of streaming models, evolving from reactive systems that respond to user queries \citep{huang2025videochatonline,Flash-VStream,xu2026streamingvlm} to proactive systems that autonomously determine when to respond as events unfold \citep{qian2025dispider,wang2026mmduet2,wang2026moss-vl,zeng2026onestreamer,yao2026joyai,li2026videochat3}. Alongside these advances, reliably tracking evolving events, determining appropriate response times, and managing historical information remain open challenges, highlighting the need for a comprehensive benchmark to evaluate proactive streaming capabilities.

\textbf{Streaming Video Understanding Benchmarks.}
Based on offline video benchmarks \citep{fang2024mmbenchvideo,wu2024longvideobench,fu2025videomme,zhou2025mlvu,wang2025lvbench,li2024mvbench,zhao2025mmvu,hong2025motionbench,vsi-bench}, existing benchmarks have significantly advanced the evaluation of streaming video understanding \citep{yang2025svbench,lin2026ostbench,RVS-Movie-bench,hu2025streamingcotbench,wang2026streameqa}. StreamingBench~\cite{lin2026streamingbench}, OVBench~\cite{huang2025videochatonline}, RTV-Bench~\cite{xun2026rtvbench}, and OVO-Bench~\cite{niu2025ovobench} primarily assess temporal understanding through reactive question answering. More recent benchmarks, including ProactiveVideoQA~\cite{wang2025proactivevideoqa} and ProReady-QA~\cite{azad2026streamready}, extend evaluation to proactive responses and their timing. SVCBench~\cite{liu2026svcbench} focuses on event counting, while RIVER~\cite{shi2026river} and OmniPro~\cite{zhao2026omnipro} explore broader streaming interaction capabilities. 
Despite these advances, existing benchmarks remain limited in task coverage and systematic evaluation across multiple event response points. Historical trajectory recovery and memory management are also insufficiently explored, while task-adaptive response windows and corresponding metrics for evaluating response timeliness remain underdeveloped.

\begin{table*}[t]
\centering
\caption{\textbf{Comparison of streaming video understanding benchmarks.}}
\label{tab:benchmark_comparison}
\setlength{\tabcolsep}{4pt}
\renewcommand{\arraystretch}{1.15}

\resizebox{\textwidth}{!}{%
\begin{tabular}{@{}lrrrrccccc@{}}
\toprule
Benchmark
& Videos
& Samples
& \shortstack{Question\\Points}
& \shortstack{Answer\\Points}
& \shortstack{Hierarchical\\Proactive Tasks}
& \shortstack{Proactive Captioning\\\& Summarization}
& \shortstack{Proactive\\Response}
& \shortstack{Mid-stream\\Proactive Activation}
& \shortstack{History Recovery\\\& Update} \\
\midrule

StreamingBench
& 900 & 4,500 & 4,500 & 4,500
& \ding{55} & \ding{55} & \ding{51} & \ding{51} & \ding{55} \\

OmniPro
& 1,262 & 2,700 & 2,700 & 9,051
& \ding{51} & \ding{55} & \ding{51} & \ding{55} & \ding{55} \\

OVBench
& 1,463 & 7,090 & 7,090 & 7,090
& \ding{55} & \ding{55} & \ding{55} & \ding{55} & \ding{55} \\

OVO-Bench
& 644 & 1,640 & 2,814 & 2,814
& \ding{55} & \ding{55} & \ding{55} & \ding{55} & \ding{55} \\

RTV-Bench
& 552 & 3,002 & 4,608 & 4,608
& \ding{55} & \ding{55} & \ding{55} & \ding{55} & \ding{55} \\

ProactiveVideoQA
& 1,377 & 1,427 & 1,427 & 3,510
& \ding{55} & \ding{55} & \ding{51} & \ding{55} & \ding{55} \\

RIVER
& 1,067 & 3,088 & 4,278 & 6,840
& \ding{55} & \ding{55} & \ding{51} & \ding{51} & \ding{55} \\

ProReady-QA
& 32 & 4,507 & 5,000 & 5,000
& \ding{55} & \ding{55} & \ding{51} & \ding{51} & \ding{55} \\

\midrule
\textbf{SVUBench (Ours)}
& \textbf{2,775}
& \textbf{3,398}
& \textbf{12,131}
& \textbf{62,881}
& \ding{51} & \ding{51} & \ding{51} & \ding{51} & \ding{51} \\
\bottomrule
\end{tabular}%
}
\end{table*}

\section{SVUBench}

\begin{figure}[t]
  \centering
  \includegraphics[width=0.95\textwidth]{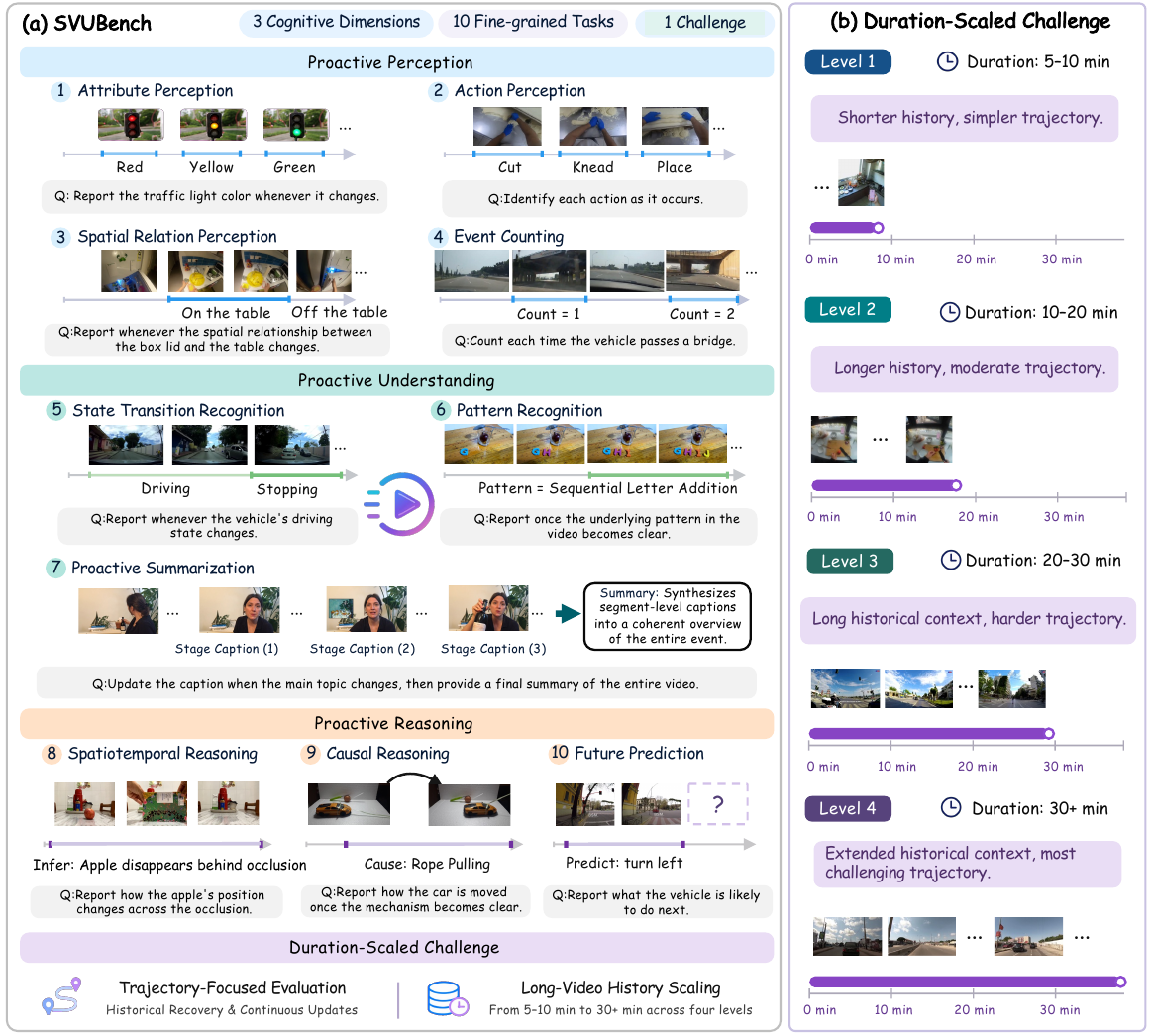}
  \caption{\textbf{Hierarchical Proactive Task Taxonomy of SVUBench.} (a) SVUBench covers ten fine-grained tasks across three cognitive dimensions and the Duration-Scaled Challenge. (b) Detailed illustration of the Duration-Scaled Challenge.}
  \label{fig:task}
\end{figure}

SVUBench comprises 12,131 questions and 62,881 answer points across ten tasks spanning three cognitive dimensions, as shown in Table~\ref{tab:benchmark_comparison} and Figure~\ref{fig:task}. It further includes a Duration-Scaled Challenge and two complementary interaction modes, Basic and Trajectory, to evaluate proactive understanding across different observation durations and task activation settings. Task-adaptive answer windows with Proactive Response Score jointly assess response content, timeliness, and historical consistency, providing a systematic evaluation of real-time perception, response control, and historical memory management. Visualizations of each task are provided in the Appendix \ref{sec:case_study_sample}.

\subsection{Hierarchical Proactive Task Taxonomy}

\textbf{Proactive Perception.}
Proactive Perception focuses on recognizing observable visual information and reporting events as they occur or finish. It encompasses four tasks, namely \textit{Attribute Perception}, \textit{Action Perception}, \textit{Spatial Relation Perception}, and \textit{Event Counting}, covering object attributes, actions, spatial relationships, and completed event instances. These tasks assess accurate visual recognition and timely responses when sufficient evidence is available.

\noindent \textbf{Proactive Understanding.}
Beyond individual observations, Proactive Understanding requires integrating temporal information to track evolving events and maintain a coherent understanding of the video. It encompasses three tasks, namely \textit{State Transition Recognition}, \textit{Pattern Recognition}, and \textit{Proactive Summarization}. The first two focus on detecting state changes and identifying temporal patterns.

Within this dimension, \textit{Proactive Summarization} further addresses information integration across local and global temporal scales. \textit{Proactive Summarization} requires models to autonomously identify meaningful event boundaries and generate captions describing salient developments in successive video segments. At the end of the video, models must further integrate these observations and their temporal relationships into a coherent final summary. This task jointly assesses proactive event segmentation, local understanding, and the ability to maintain and synthesize information throughout the entire video.

\noindent \textbf{Proactive Reasoning.}
Proactive Reasoning further emphasizes drawing conclusions from multiple visual cues and their underlying relationships. It consists of three tasks, namely \textit{Spatiotemporal Reasoning}, \textit{Causal Reasoning}, and \textit{Future Prediction}. Spatiotemporal Reasoning involves interpreting spatial and temporal relationships between events, while Causal Reasoning examines their cause and effect connections. Future Prediction requires anticipating upcoming events based on observed cues and contextual evidence. Collectively, these tasks assess the ability to construct well-supported inferences.
% and determine when the available evidence justifies a response.

\noindent \textbf{Duration-Scaled Challenge.}
This challenge evaluates proactive understanding and memory management over extended video streams. It organizes long videos into four duration tiers, covering selected tasks from our taxonomy. As observation histories grow, models must retain and retrieve relevant information, integrate new evidence, and continuously update their responses. Evaluation across these tiers reveals how increasing video duration affects memory management, response timeliness, and answer quality.

\subsection{Dual Mode Interaction}

To evaluate proactive understanding under different interaction scenarios, we introduce two complementary modes.

\noindent \textbf{Basic.}
The task is activated at video onset and remains active throughout the stream. Models continuously monitor relevant events and proactively provide the \emph{Current Answer} within the valid response window once sufficient evidence becomes available. As new events emerge, models must update their responses accordingly. This mode evaluates sustained event monitoring, response timeliness, and answer accuracy during continuous observation.

\noindent \textbf{Trajectory.}
In Trajectory mode, the task is introduced midway through the video, requiring models to retrieve relevant information from previously observed content. Within the first valid response window, models must provide the \emph{Current Answer} and a \emph{Historical Trace} that reconstructs all relevant past events and their timestamps. Subsequent responses include the updated \emph{Current Answer} and \emph{Last Evidence}, identifying the event and timestamp associated with the previous response. This mode   evaluates models' ability to manage historical memory, accurately retrieve past evidence, reconstruct event trajectories, and update their understanding as new evidence emerges.

\subsection{Dataset Construction}

\begin{figure}[t]
  \centering
  \includegraphics[width=1\linewidth]{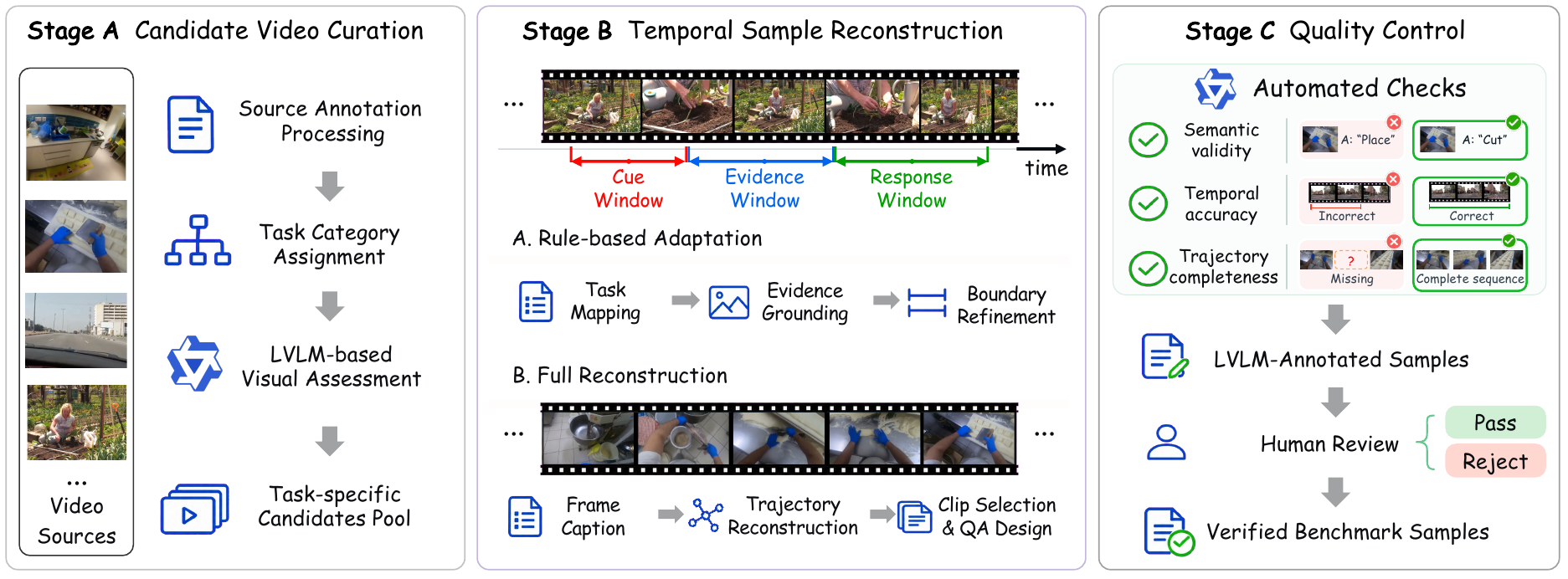}
  \caption{SVUBench construction pipeline.}
  \label{fig:pipeline}
\end{figure}

We construct SVUBench through three stages, including candidate video curation, temporal sample reconstruction, and quality control, as illustrated in Figure~\ref{fig:pipeline}. The pipeline integrates model-assisted screening and annotation with human verification to ensure semantic validity, temporal accuracy, and annotation reliability. Details and prompts are provided in Appendix \ref{app:prompts}.

\noindent \textbf{Candidate Video Curation.}
We collect candidate videos from VidOR~\cite{vidor}, Perception Test~\cite{perceptiontest}, Vript~\cite{yang2024vript}, and Ego4D~\cite{grauman2022ego4d}. Rule-based adapters first standardize source annotations. Guided by these annotations and our task taxonomy, Qwen3.8-27B~\citep{team2026qwen3-8} performs initial task classification using sparsely sampled frames. Unclassified candidates are examined by Qwen3.5-35B-A3B~\citep{team2026qwen3-5} with denser frame sampling. Successfully classified videos are assigned to task-specific candidate pools, while the remaining videos are discarded. Long videos for the Duration-Scaled Challenge are separately curated into four duration tiers.

\noindent \textbf{Temporal Sample Reconstruction.}
We employ two annotation strategies depending on the available source annotations. Rule-based adaptation maps existing annotations to proactive tasks, grounds supporting evidence, and refines temporal boundaries. Otherwise, frame captions guide event trajectory reconstruction, followed by clip selection and task-specific QA generation. We also define task activation times for both interaction modes. Basic activates tasks at clip onset, while Trajectory places the first query midway between clip onset and the first reportable event. Up to four additional queries are placed near 30\%, 50\%, 70\%, and 85\% of the clip duration, provided they fall between a historical event and a future reportable event. 

\noindent \textbf{Quality Control.}
All MLLM-annotated samples undergo automated checks for semantic validity, temporal accuracy, and trajectory completeness. Human reviewers further verify event coverage, refine temporal boundaries, and ensure consistency between answers and supporting visual evidence. Samples failing verification are corrected or rejected, yielding the final verified benchmark.

\subsection{Dataset Statistics}

\begin{figure}[t]
  \centering
  \includegraphics[width=1\linewidth]{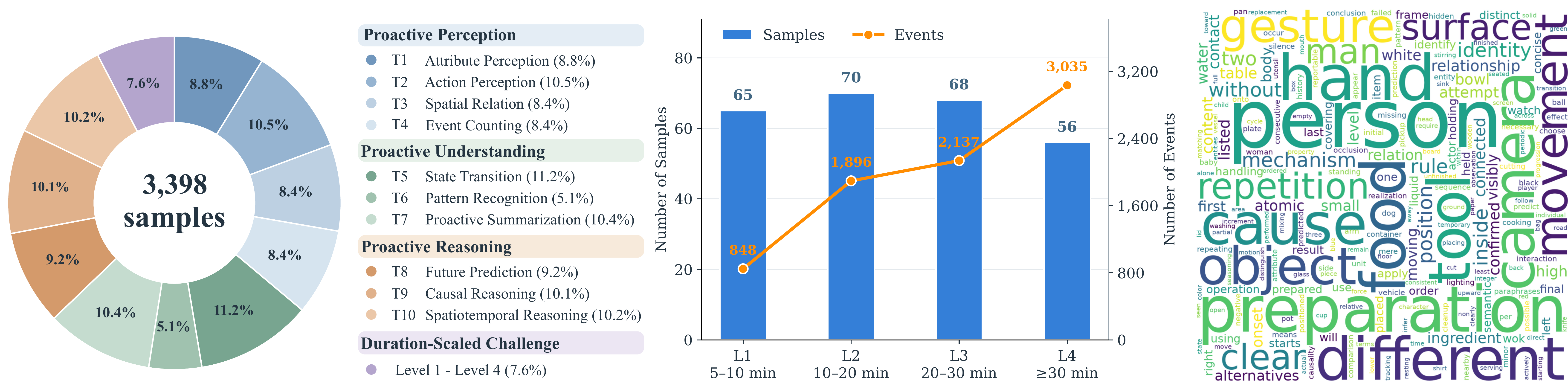}
  \caption{\textbf{Dataset statistics of SVUBench.} Left: Task Distribution. Middle: Duration-Scaled Challenge. Right: Linguistic Characteristics.}
  \label{fig:statistics}
\end{figure}

SVUBench comprises 3,398 samples spanning ten proactive tasks across Perception, Understanding, and Reasoning. As shown in Figure \ref{fig:statistics}, individual categories account for 5.1\%–11.2\% of all samples, providing broad coverage of visual recognition, temporal understanding, and higher-level reasoning. The 
The Duration-Scaled Challenge further includes 259 samples covering tasks T1, T2, T3, T4, T5, T7, and T8 across four duration tiers, ranging from 5--10 minutes to over 30 minutes, with 7,916 annotated events in total. These extended event sequences support evaluation of sustained monitoring, historical retrieval, and continuous response updates. The query word cloud illustrates diverse linguistic and semantic content. Together with its broad task coverage and extensive event annotations, this diversity makes SVUBench a comprehensive benchmark for proactive streaming video understanding.

\begin{figure}[t]
  \centering
  \includegraphics[width=1\linewidth]{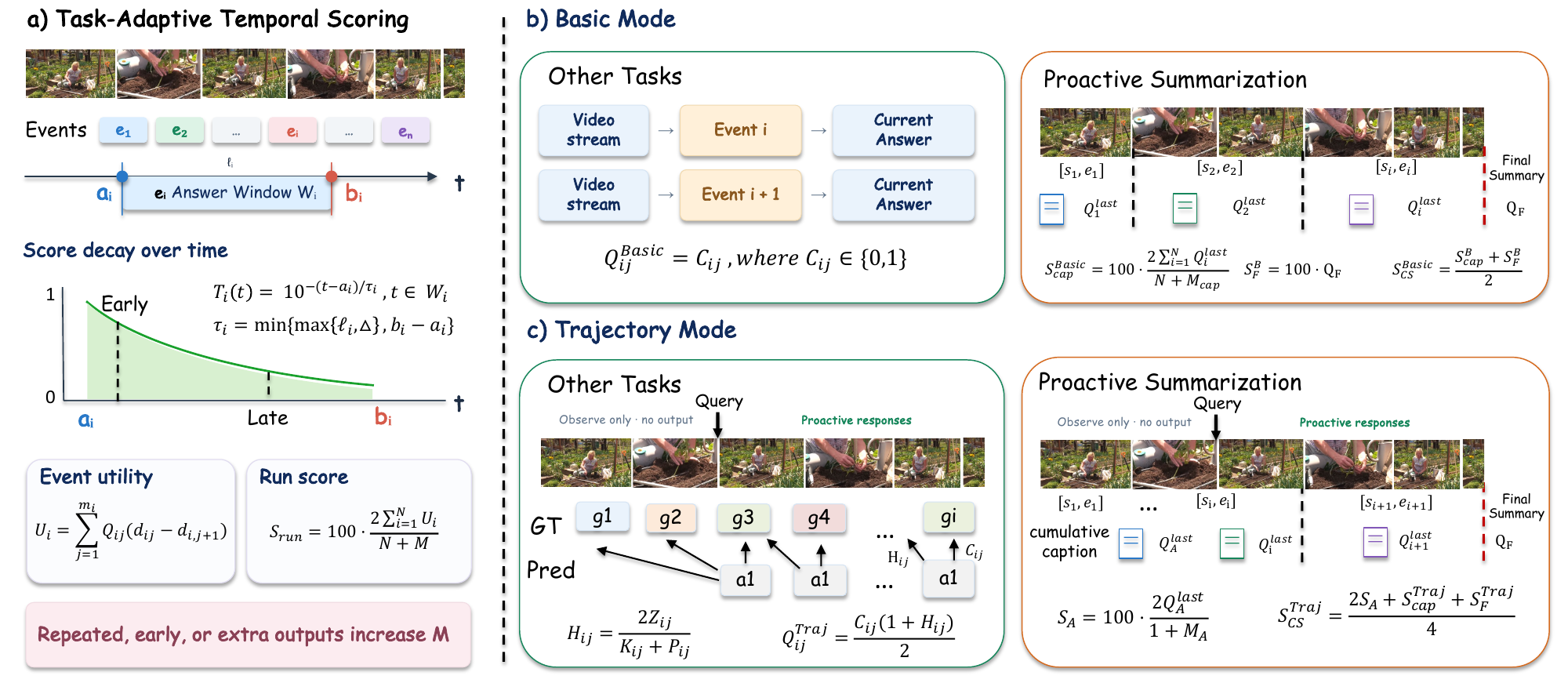}
  \caption{\textbf{Proactive Response Score.} (a) Task-adaptive temporal scoring with event-specific response windows, time-decaying rewards, and penalties for redundant outputs. (b) Basic mode evaluation of event-based tasks and proactive summarization. (c) Trajectory mode evaluation incorporating historical recovery, current answer quality, cumulative captioning, and continuous evidence updates.}
  \label{fig:metric}
  % \vspace{-3mm}
\end{figure}

\subsection{Proactive Response Evaluation}

\noindent We introduce the \emph{Proactive Response Score} (PRS) to evaluate response correctness, timeliness, and historical consistency, as shown in Figure \ref{fig:metric}. The framework incorporates task-specific response windows, content quality, and penalties for delay, missed or redundant responses. Evaluation prompts and detailed response window definitions for each task are provided in Appendix \ref{supp_prompts_eval} and Appendix \ref{app:answer_windows}.

\noindent \textbf{Task-Adaptive Temporal Scoring.}
For each target event $i$, we define a valid response window $W_i$ with start $a_i$ and deadline $b_i$, determined by when sufficient evidence available and when the response remains relevant. Earlier responses within the window receive higher temporal scores
\begin{equation}
T_i(t)=
\begin{cases}
10^{-(t-a_i)/\tau_i}, & t\in W_i,\\
0, & \text{otherwise},
\end{cases}
\end{equation}
where $\tau_i=\min\{\max(\ell_i,\Delta),b_i-a_i\}$, $\ell_i$ is the local event duration, and $\Delta$ is the video sampling interval. Response time is determined by the last observed frame when generation begins. To account for discrete response opportunities, the first available response slot within each valid window receives no temporal penalty. We combine response quality and temporal scores to compute event utility and the normalized run score
\begin{equation}
U_i=\sum_{j=1}^{m_i}Q_{ij}(d_{ij}-d_{i,j+1}),
S_{\mathrm{run}}=100\cdot\frac{2\sum_{i=1}^{N}U_i}{N+M}.
\end{equation}
Here, $m_i$ is the number of responses within $W_i$, $Q_{ij}$ denotes response quality, $d_{ij}=T_i(t_{ij})$, and $d_{i,m_i+1}=0$. $N$ and $M$ denote the numbers of reference events and submitted responses. This formulation rewards accurate and timely responses while penalizing missed events and redundant outputs. All submitted responses, including incorrect, repeated, and out-of-window outputs, contribute to $M$.

\noindent \textbf{Basic Mode Scoring.}
In Basic mode, response quality depends on the correctness of the current answer, defined as $Q_{ij}^{\mathrm B}=C_{ij}$, where $C_{ij}\in\{0,1\}$. The resulting event utilities are aggregated using the normalized scoring formulation above, measuring both answer accuracy and response timeliness throughout continuous observation.

For Proactive Summarization, models generate a \emph{Current Caption} for each semantic segment and a \emph{Final Summary} after video completion. A fixed judging prompt assigns semantic quality scores in $\{0,0.25,0.5,0.75,1\}$. Each segment is evaluated using its last complete caption within the corresponding segment window. Given $n$ reference segments and $M_{\mathrm{cap}}$ caption submissions, the scores are
\begin{equation}
S_{\mathrm{cap}}^{\mathrm Basic}
=100\cdot\frac{2\sum_{i=1}^{n}Q_i^{\mathrm{last}}}
{n+M_{\mathrm{cap}}},
\qquad
S_F^{\mathrm Basic}=100\cdot Q_F,
\end{equation}
where $Q_F$ denotes the semantic quality of the final summary. The Basic summarization score equally weights $S_{\mathrm{cap}}^{\mathrm Basic}$ and $S_F^{\mathrm Basic}$ as
$S_{\mathrm{CS}}^{\mathrm Basic}=(S_{\mathrm{cap}}^{\mathrm Basic}+S_F^{\mathrm Basic})/2$.

\noindent \textbf{Trajectory Mode Scoring.}
Trajectory mode further evaluates historical recovery and continuous evidence updates. The first response after task activation must provide a \emph{Historical Trace} containing all relevant past events and their timestamps, while subsequent responses include \emph{Last Evidence} referencing the event and timestamp associated with the previous response. For the $j$-th response to event $i$, let $K_{ij}$ and $P_{ij}$ denote the numbers of reference and predicted historical events, respectively. We perform order-preserving, one-to-one matching, with $Z_{ij}$ representing the maximum total matching credit. Semantically correct matches receive full credit if their timestamps fall within the reference evidence windows, and half credit otherwise. Historical quality and response quality are computed as
\begin{equation}
H_{ij}=\frac{2Z_{ij}}{K_{ij}+P_{ij}},
\qquad
Q_{ij}^{\mathrm{Traj}}
=C_{ij}\frac{1+H_{ij}}{2}.
\end{equation}
We set $H_{ij}=1, \text{if } K_{ij}+P_{ij}=0.$ This formulation rewards accurate historical reconstruction while retaining partial credit for correct current answers. Event utilities follow the same temporal scoring and response penalties as Basic.

For \emph{Proactive Summarization}, the first valid response after task activation jointly provides a \emph{Current Caption} for the current segment and a \emph{Cumulative Caption} covering all segments observed so far. Subsequent responses provide segment captions, followed by a \emph{Final Summary} at video completion. The Cumulative Caption is evaluated using the same semantic judging protocol without requiring timestamps. Let $Q_A^{\mathrm{last}}$ denote its quality in the last complete submission within the first response window, and $M_A$ the number of submissions in that window. The cumulative caption score and overall Trajectory summarization score are
\begin{equation}
S_A=100\cdot\frac{2Q_A^{\mathrm{last}}}{1+M_A},
S_{\mathrm{CS}}^{\mathrm{Traj}}
=\frac{2S_A+S_{\mathrm{cap}}^{\mathrm{Traj}}+S_F^{\mathrm{Traj}}}{4}.
\end{equation}
Here, $S_{\mathrm{cap}}^{\mathrm{Traj}}$ evaluates current captions from the first valid segment onward, while $S_F^{\mathrm{Traj}}$ evaluates the final summary.

\section{Experiments}

\noindent\textbf{Evaluation Models.} We evaluate 17 representative models covering four categories of Video MLLMs. \textit{Offline Models} include InternVL3.5-8B \citep{wang2025internvl3.5}, Qwen3-VL-8B-Instruct \citep{bai2025qwen3}, Qwen3.5-9B \citep{team2026qwen3-5}, Qwen3.5-35B-A3B \citep{team2026qwen3-5}, and Qwen3.8-27B \citep{team2026qwen3-8}. \textit{Reactive Streaming Models} include StreamingVLM \citep{xu2026streamingvlm}, VideoChat-Online \citep{huang2025videochatonline},  FlashVStream-7B \citep{Flash-VStream}, and StreamForest \citep{zeng2026streamforest}, which support incremental video processing and query-driven interaction. \textit{Proactive Streaming Models} include VideoLLM-online \citep{VideoLLM-online}, MMDuet2-3B \citep{wang2026mmduet2}, Dispider \cite{qian2025dispider}, MOSS-VL-Realtime-11B \citep{wang2026moss-vl}, OneStreamer-4B \citep{zeng2026onestreamer}, JoyAI-VL-Interaction \citep{yao2026joyai}, and VideoChat3 \citep{li2026videochat3}, which further explore autonomous response generation during continuous observation. We include DeepSeek-v4.1-Flash \citep{xu2026deepseek-v4-1-flash} as the API baseline. These models cover different architectures and interaction paradigms, enabling a comprehensive comparison of video understanding capabilities.

\noindent\textbf{Implementation Details.}
All models follow a unified evaluation protocol, with videos sampled at 2 FPS and one response opportunity per second. We use publicly released codebases and checkpoints, retaining each model's default inference settings and specialized mechanisms. For offline and reactive models, we adopt prefix-based evaluation, providing only observed video frames, the task query, and previously generated responses at each response interval. Proactive models process frames incrementally and respond through their native mechanisms. We standardize task prompts, output formats, and evaluation constraints across models. All experiments are conducted on 128 NVIDIA H20 GPUs. More details are provided in Appendix \ref{app:proactive_models}.

\begin{table*}[t]
\centering
\caption{\textbf{Performance comparison across ten proactive tasks
on SVUBench under Basic and Trajectory modes.}
Each cell reports Basic / Trajectory scores.
The best score in each task column and the Avg. column
is shown in bold separately for each mode.}
\label{tab:main_results}
\label{tab:trajectory_results}

\setlength{\tabcolsep}{2pt}
\renewcommand{\arraystretch}{1.13}

\resizebox{\textwidth}{!}{%
\begin{tabular}{@{}>{\bfseries}l*{11}{c}@{}}
\toprule
& \multicolumn{4}{c}{\textbf{Proactive Perception}}
& \multicolumn{3}{c}{\textbf{Proactive Understanding}}
& \multicolumn{3}{c}{\textbf{Proactive Reasoning}}
& \\
\cmidrule(lr){2-5}
\cmidrule(lr){6-8}
\cmidrule(lr){9-11}
Models
& \textbf{T1} & \textbf{T2} & \textbf{T3} & \textbf{T4}
& \textbf{T5} & \textbf{T6} & \textbf{T7}
& \textbf{T8} & \textbf{T9} & \textbf{T10}
& \textbf{Avg.} \\
\midrule

\multicolumn{12}{c}{\textit{API Model}} \\
\midrule
DeepSeek-v4.1-Flash
& 4.64/2.97 & 10.08/5.63 & 7.11/4.94 & 2.90/1.85
& 8.08/4.77 & 4.06/3.24 & 34.80/\textbf{37.83}
& 2.24/1.63 & 3.98/3.77 & 2.81/1.89 & 8.07/6.85 \\
\midrule

\multicolumn{12}{c}{\textit{Offline Models}} \\
\midrule
InternVL3.5-8B
& 1.45/1.26 & 3.46/2.52 & 2.92/2.29 & 1.42/0.70
& 3.05/1.91 & 2.46/3.07 & 14.13/9.96
& 1.06/0.78 & 1.93/3.13 & 1.49/2.17 & 3.34/2.78 \\

Qwen3-VL-8B-Instruct
& 3.60/2.24 & 8.22/4.74 & 5.97/3.83 & 2.58/1.48
& 5.52/3.27 & 3.12/3.00 & 34.01/22.87
& 2.09/1.50 & 3.40/3.92 & 2.19/1.41 & 7.07/4.83 \\

Qwen3.5-9B
& 4.00/2.32 & 8.67/5.20 & 6.31/3.94 & 2.84/1.55
& 6.27/3.67 & 3.32/3.17 & 32.56/23.34
& 2.17/1.49 & 3.75/3.36 & 2.14/1.76 & 7.20/4.98 \\

Qwen3.5-35B-A3B
& 4.41/2.54 & 9.78/5.20 & 6.62/4.14 & 4.40/2.09
& 6.42/3.56 & 3.22/3.21 & \textbf{40.38}/23.10
& 2.25/1.77 & 3.74/\textbf{4.06} & 2.65/1.42
& 8.39/5.11 \\

Qwen3.8-27B
& 4.35/2.41 & 8.78/4.79 & 7.01/4.28 & 3.99/2.11
& 5.79/3.46 & 3.33/3.47 & 36.26/37.76
& 2.41/1.62 & 3.46/3.96 & 2.49/1.78 & 7.79/6.56 \\
\midrule

\multicolumn{12}{c}{\textit{Reactive Streaming Models}} \\
\midrule
StreamingVLM
& 2.13/0.35 & 1.41/0.04 & 0.57/0.71 & 0.65/0.59
& 0.68/0.00 & 1.30/0.00 & 7.01/2.10
& 0.00/0.00 & 0.49/0.16 & 0.58/0.00 & 1.48/0.39 \\

VideoChat-Online
& 3.57/0.61 & 2.47/0.58 & 4.74/1.06 & 3.42/1.41
& 1.05/0.50 & 2.06/0.57 & 15.35/7.47
& 0.00/0.00 & 1.78/0.25 & 1.80/0.06 & 3.62/1.25 \\

FlashVStream-7B
& 1.82/0.87 & 3.64/1.55 & 4.53/1.74 & 2.42/0.91
& 2.92/1.37 & 1.84/2.05 & 17.56/8.70
& 1.18/0.71 & 2.73/2.30 & 0.90/1.20 & 3.95/2.14 \\

StreamForest
& 0.22/0.58 & 0.08/0.44 & 2.50/0.56 & 0.76/0.60
& 0.16/0.08 & 0.57/0.00 & 16.02/8.06
& 0.00/0.00 & 0.19/0.23 & 0.43/0.23 & 2.09/1.08 \\
\midrule

\multicolumn{12}{c}{\textit{Proactive Streaming Models}} \\
\midrule
VideoLLM-online
& 0.65/0.17 & 2.00/1.06 & 0.06/0.07 & 0.34/0.95
& 0.74/0.56 & 1.46/0.33 & 13.01/6.51
& 0.55/0.01 & 0.37/0.25 & 1.07/0.31 & 2.02/1.02 \\

MMDuet2-3B
& 0.21/0.26 & 1.37/1.72 & 0.47/0.29 & 0.13/0.03
& 3.21/2.06 & 1.22/0.00 & 20.31/10.25
& 0.77/0.44 & 0.32/0.26 & 1.47/0.01 & 2.95/1.53 \\

Dispider
& 8.13/3.87 & 1.41/1.13 & 2.05/3.68 & 7.36/1.35
& 4.00/3.87 & \textbf{4.46}/1.38 & 16.87/2.59
& 0.31/0.54 & 1.05/0.58 & 4.97/0.49 & 5.06/1.95 \\

MOSS-VL-Realtime-11B
& 23.08/\textbf{14.08}
& \textbf{29.38}/\textbf{12.39}
& 16.65/\textbf{9.77}
& 5.02/2.06
& \textbf{32.64}/\textbf{15.86}
& 2.54/3.60
& 27.75/27.55
& \textbf{5.60}/\textbf{3.18}
& 0.55/2.64
& 2.33/0.99
& \textbf{14.55}/\textbf{9.21} \\

OneStreamer-4B
& \textbf{28.87}/10.44
& 14.61/2.93
& \textbf{19.35}/6.23
& \textbf{10.60}/\textbf{5.34}
& 20.37/4.05
& 1.49/0.20
& 18.25/3.20
& 2.98/1.20
& \textbf{6.10}/0.85
& 10.20/0.91
& 13.28/3.53 \\

JoyAI-VL-Interaction
& 19.03/7.12 & 21.89/11.05 & 10.74/6.86 & 6.04/2.80
& 29.34/12.35 & 1.02/\textbf{8.85} & 32.36/32.26
& 1.73/2.41 & 1.09/3.33 & 3.31/3.51 & 12.66/9.05 \\

VideoChat3
& 22.62/11.06 & 13.52/5.86 & 14.80/6.89 & 6.76/2.34
& 16.52/8.27 & 2.41/2.42 & 13.15/8.56
& 3.73/1.34 & 5.61/3.87
& \textbf{12.60}/\textbf{4.33} & 11.17/5.50 \\
\bottomrule
\end{tabular}%
}

\vspace{4pt}
\parbox{\textwidth}{%
\footnotesize
\textbf{Tasks:}
T1: Attribute Perception;
T2: Action Perception;
T3: Spatial Relation Perception;
T4: Event Counting;
T5: State Transition Recognition;
T6: Temporal Pattern Recognition;
T7: Proactive Summarization;
T8: Future Prediction;
T9: Causal Reasoning;
T10: Spatiotemporal Reasoning.
}
\vspace{-3mm}
\end{table*}

\subsection{Basic-Mode Results and Analysis}

Table~\ref{tab:main_results} highlights the limitations of offline and reactive models under proactive evaluation. Frequent response opportunities lead to substantial redundant and out-of-window outputs, resulting in considerable penalties. Some reactive models produce few valid responses, potentially due to mismatches between their training distributions and the evaluation protocol. Despite their strong underlying video understanding capabilities, these models lack effective response control for proactive streaming interaction.

\begin{figure*}[t]
\centering

% Left half: table
\begin{minipage}[t]{0.49\textwidth}
\vspace{0pt}
\centering
\captionof{table}{Results on the Duration-Scaled Challenge.
Each entry reports Basic / Trajectory scores.}
\label{tab:duration_scaled_results}
\small
\setlength{\tabcolsep}{5pt}
\renewcommand{\arraystretch}{3}
\resizebox{\linewidth}{!}{%
\begin{tabular}{lccccc}
\toprule
\textbf{Model}
& \textbf{L1}
& \textbf{L2}
& \textbf{L3}
& \textbf{L4}
& \textbf{Average} \\
\midrule
MOSS-VL-Realtime
& \textbf{19.59 / 11.54}
& 10.11 / 6.61
& 8.96 / \textbf{5.97}
& \textbf{7.25 / 3.60}
& \textbf{11.69 / 7.12} \\

OneStreamer
& 14.08 / 4.84
& \textbf{10.76} / 4.31
& \textbf{9.84} / 4.47
& 5.45 / 2.42
& 10.45 / 4.22 \\

JoyAI-VL-Interaction
& 11.47 / 8.49
& 8.93 / \textbf{6.67}
& 6.95 / 5.43
& 4.70 / 2.53
& 7.64 / 5.57 \\

VideoChat3
& 14.09 / 6.80
& 8.56 / 3.54
& 6.51 / 2.86
& 3.80 / 2.31
& 6.87 / 3.45 \\
\bottomrule
\end{tabular}%
}
\end{minipage}%
\hfill
% Right half: caption above the two plots
\begin{minipage}[t]{0.49\textwidth}
\vspace{0pt}
\centering
\captionsetup{position=top}
\captionof{figure}{\textbf{Failure mode analysis.}
Basic evaluation (left) and Trajectory evaluation (right).}
\label{fig:mode_analysis}

\begin{minipage}[t]{0.49\linewidth}
\vspace{0pt}
\centering
\includegraphics[width=\linewidth]{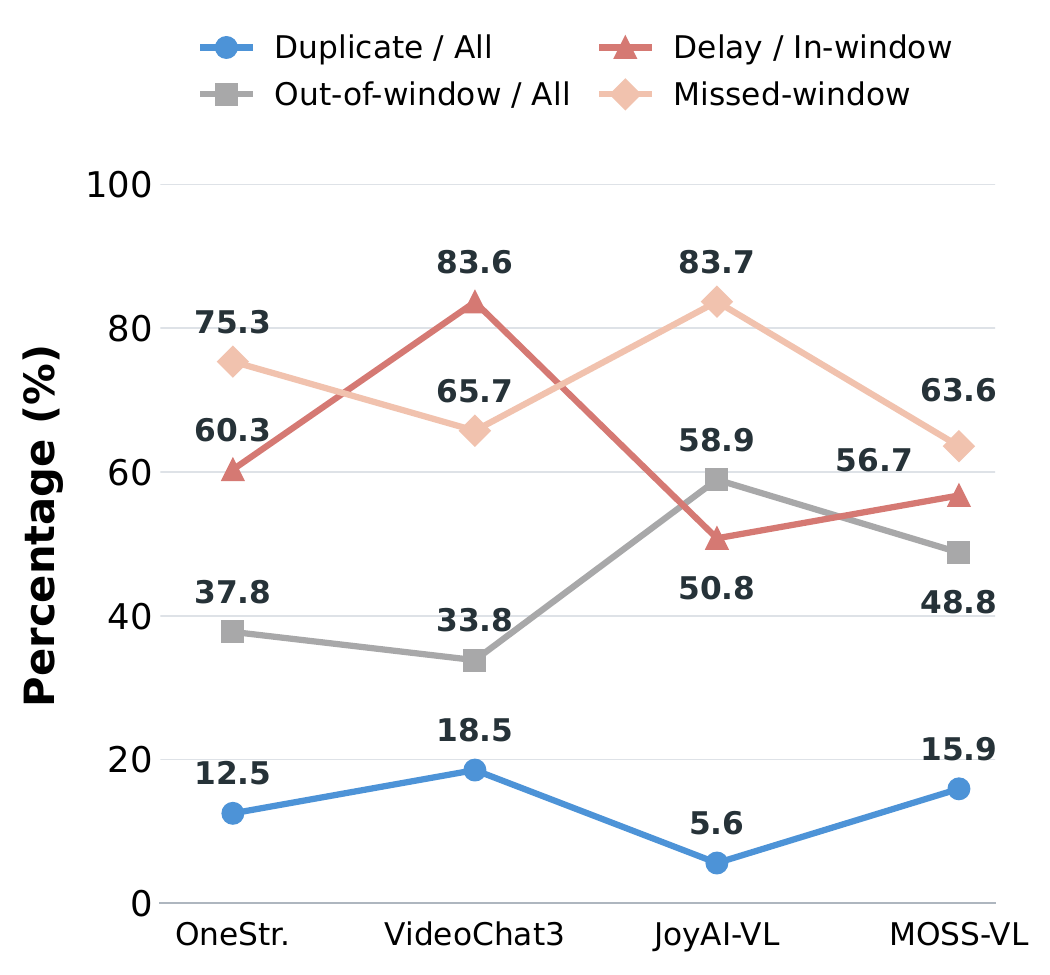}
\end{minipage}%
\hfill
\begin{minipage}[t]{0.49\linewidth}
\vspace{0pt}
\centering
\includegraphics[width=\linewidth]{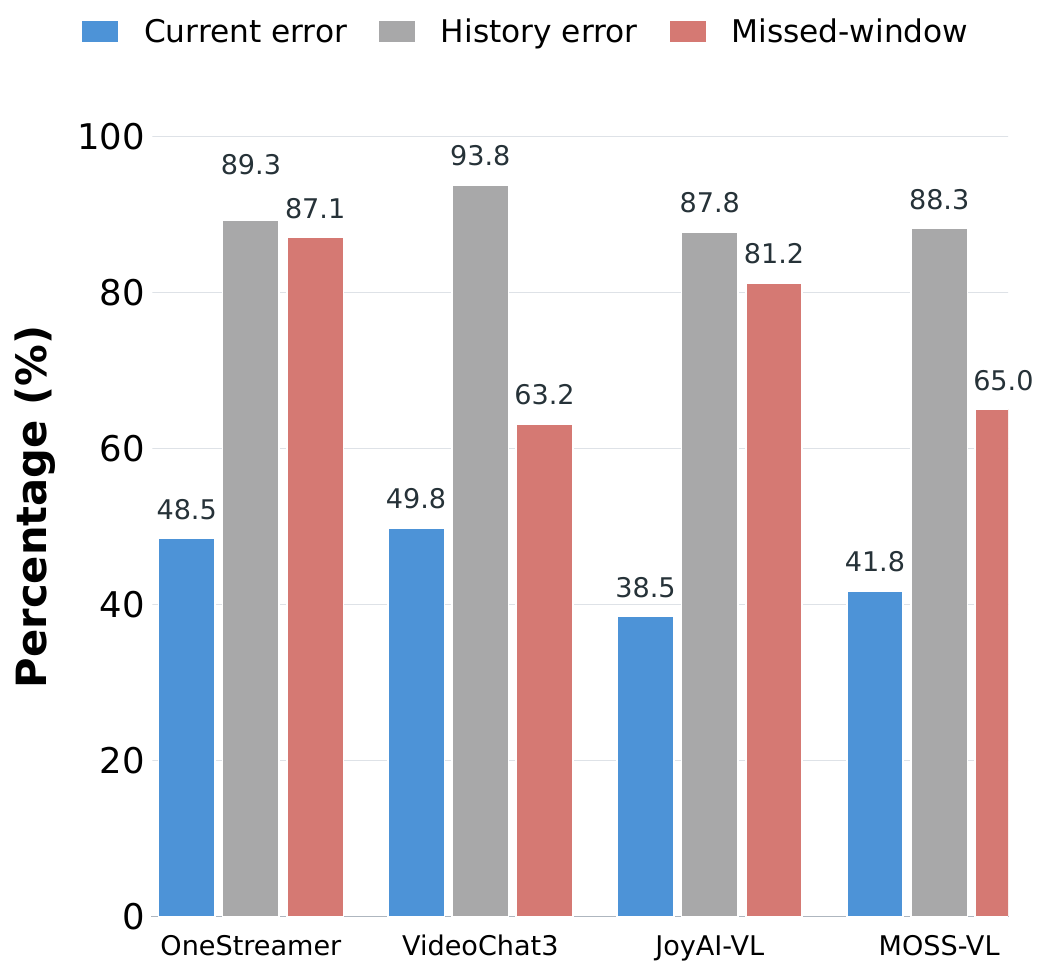}
\end{minipage}
\end{minipage}

\end{figure*}

Proactive models achieve stronger overall performance, with MOSS leading at 14.55. However, performance varies considerably across tasks, particularly in temporal pattern recognition and causal reasoning, suggesting uneven capabilities and room for more balanced training. Figure~\ref{fig:mode_analysis} further exposes limitations in response control. Models frequently repeat previously reported events, generating redundant outputs, while responses outside valid windows reveal difficulties in determining appropriate response timing. Even within valid windows, substantial delays indicate an inability to respond promptly once sufficient evidence becomes available. Moreover, missed-window rates of 63.6\% to 83.7\% highlight persistent failures in detecting and responding to relevant events. These findings reveal considerable room for improvement in task competence, continuous event awareness, response timing, and output control. Extensive qualitative case studies are in Appendix \ref{sec:case_study}.

\subsection{Trajectory-Mode Results and Analysis}

Table~\ref{tab:trajectory_results} shows further performance declines for offline and reactive models under Trajectory mode. This further exposes their limitations in memory management, historical retrieval, and sustained proactive interaction.

Among proactive models, MOSS and JoyAI achieve the highest average scores of 9.21 and 9.05, respectively, but substantial challenges remain. The response control issues observed in Basic mode persist, while additional historical reporting requirements further expose weaknesses in instruction following and memory management. As shown in Figure~\ref{fig:mode_analysis}, historical field error rates reach 87.8\% to 93.8\%, reducing scores even when current answers are correct. These findings suggest that existing proactive models primarily emphasize perceiving new information and generating responses, with insufficient attention to preserving and retrieving historical evidence. Despite advances in continuous visual perception, effective memory management and historical information integration remain important directions for improving proactive streaming understanding.

\subsection{Duration-Scaled Challenge Results}
% \vspace{-1mm}

As video duration increases from L1 to L4, evaluated models exhibit a progressive decline in performance in Table \ref{tab:duration_scaled_results}. This trend highlights how current streaming models struggle with longer videos and denser events, particularly in managing accumulated information, leveraging historical evidence, and maintaining reliable real-time perception. Detailed results and analysis are in Appendix \ref{app:detailed_duration}. 

\begin{table}[t]
\centering
\caption{Diagnostic evaluation of content quality and timing.}
\label{tab:diagnostic_evaluation}
\small
\setlength{\tabcolsep}{3pt}
\renewcommand{\arraystretch}{1}
\resizebox{\linewidth}{!}{%
\begin{tabular}{@{}l*{8}{c}@{}}
\toprule
& \multicolumn{4}{c}{\textbf{Basic}}
& \multicolumn{4}{c}{\textbf{Trajectory}} \\
\cmidrule(lr){2-5}
\cmidrule(lr){6-9}
\textbf{Model}
& \textbf{Full}
& \textbf{Content}
& \textbf{Timing}
& \textbf{w/o Delay}
& \textbf{Full}
& \textbf{Content}
& \textbf{Timing}
& \textbf{w/o Delay} \\
\midrule
DeepSeek-v4.1-Flash
& 8.07 & \textbf{63.00} & 13.02 & 13.24
& 6.85 & \textbf{69.99} & 14.90 & 14.96 \\
JoyAI-VL-Interaction
& 12.66 & 21.37 & 20.06 & 23.51
& 9.05 & 17.97 & 24.83 & 31.93 \\
MOSS-VL-Realtime
& \textbf{14.55} & 22.28 & \textbf{24.18} & \textbf{30.54}
& \textbf{9.21} & 25.93 & \textbf{25.00} & \textbf{32.81} \\
\bottomrule
\end{tabular}%
}
\vspace{-3mm}
\end{table}

\subsection{Performance Gap Analysis}
% \vspace{-1mm}

To analyze the gap to proactive streaming understanding, we rescore DeepSeek, MOSS, and JoyAI under three diagnostic settings to disentangle content quality from response timing in Table~\ref{tab:diagnostic_evaluation}. The Content score removes temporal and redundancy penalties, while the Timing score assumes correct content, with an additional variant excluding response delays. DeepSeek demonstrates substantially stronger content understanding but suffers from poor response control, whereas MOSS and JoyAI better regulate their outputs but exhibit limited content accuracy. Removing delay penalties further reveals considerable response latency in proactive models. Moreover, their limited historical reporting under Trajectory highlights weaknesses in memory management. These findings highlight the need to enhance real-time perception during proactive model training while balancing response timeliness and strengthening memory management. Detailed results are provided in Appendix \ref{app:detailed_gap_analysis}.

\section{Conclusion and Implications}
% \vspace{-1mm}

We introduce SVUBench, a comprehensive benchmark for proactive streaming video understanding, covering ten tasks across Perception, Understanding, and Reasoning, including proactive captioning and final summarization. Its complementary Basic and Trajectory modes evaluate sustained proactive responses, historical recovery, and continuous evidence updates. Proactive Response Score with designed answer windows measures response correctness, timeliness, and historical consistency. Extensive experiments reveal substantial limitations in real-time perception, response timing, and memory management, highlighting the need for more effective event monitoring, timely responses, and historical information retrieval. By systematically identifying these challenges, SVUBench provides a foundation for advancing reliable proactive streaming video understanding.

\clearpage

\bibliographystyle{plainnat}
\bibliography{references}

\clearpage

\clearpage
\setcounter{page}{1}
% \maketitlesupplementary

% \clearpage
% \setcounter{page}{1}
% \maketitlesupplementary

\renewcommand{\thesection}{\Alph{section}}
\setcounter{section}{0}

\makeatletter
\renewcommand{\theHsection}{\thesection}%
\renewcommand{\theHsubsection}{\thesection.\arabic{subsection}}%
\makeatother

\section{Appendix Overview}

The appendix is organized as follows:
\begin{itemize}
\setlength{\itemsep}{0pt}
\setlength{\parskip}{0pt}
\item[] Sec.~\ref{app:license}:
License
\item[] Sec.~\ref{app:prompts}:
Prompts for Data Construction
\item[] Sec.~\ref{supp_prompts_eval}:
Prompts for Evaluation
\item[] Sec.~\ref{app:answer_windows}:
Task-Adaptive Answer Windows
\item[] Sec.~\ref{app:proactive_models}:
Details of Proactive Video MLLMs
\item[] Sec.~\ref{supp_additional_results}:
Additional Results
\item[] Sec.~\ref{supp_human}:
Human Annotation Ethics and Procedures
\item[] Sec.~\ref{supp_limitation}:
Limitations and Future Work
\item[] Sec.~\ref{sec:case_study_sample}:
Qualitative Examples of SVUBench
\item[] Sec.~\ref{sec:case_study}:
Case Study
\end{itemize}

\section{License}
\label{app:license}

\noindent\textbf{Source Datasets.}
SVUBench is developed for academic research based on videos and source annotations from VidOR~\cite{vidor}, Perception Test~\cite{perceptiontest}, Vript~\cite{yang2024vript}, and Ego4D~\cite{grauman2022ego4d}.
VidOR annotations are released under CC BY-SA 4.0, while the underlying YFCC100M videos remain subject to the licenses of individual media items.
Perception Test is released under CC BY 4.0.
Vript is subject to academic-research-only terms, which prohibit commercial use and restrict the redistribution and modification of source content.
Ego4D is governed by its license agreement, which requires users to accept the terms and obtain authorization before accessing the data.

\noindent\textbf{Use and Redistribution.}
All third-party videos and source annotations remain the pr operty of their respective rights holders and are subject to their original licenses and terms of use.
SVUBench's task-specific annotations do not supersede or override these upstream terms.
Any use or redistribution of source-derived materials must comply with the applicable attribution, ShareAlike, research-use, privacy, and access requirements.
Where redistribution is restricted, users must obtain the underlying videos directly from the original providers under the corresponding licenses or agreements.
We will release SVUBench under the Creative Commons Attribution 4.0 International (CC BY 4.0) license and the MIT License.

\section{Prompts for Data Construction}
\label{app:prompts}

This section presents the prompts used for video curation, sample construction, task-specific reporting, and evaluation. Angle-bracketed fields denote placeholders populated with sample-specific inputs. Construction prompts operate on supplied frames and annotations, whereas evaluation prompts restrict models to the video prefix available at each observation time.

% ============================================================
% 1. Video curation
% ============================================================

\subsection{Video Curation and Task Assignment.} 

The curation prompt identifies the proactive tasks supported by a candidate video. It requires visual evidence for the necessary temporal structure and marks cases with insufficient observations as uncertain.

\begin{svuprompt}{Video Curation and Task Assignment}
You are curating videos for a streaming video understanding benchmark.

Inspect the supplied video frames in chronological order and determine which tasks the video can support. A video may support multiple tasks. Base your decisions on visible evidence; use source annotations only as retrieval hints.

Task definitions:
1. Attribute perception: A directly observable property of the same entity changes over time.
2. Event counting: Independent, successfully completed instances of a clearly defined event can be counted.
3. Action perception: The same actor transitions between distinguishable main atomic actions.
4. Spatial relation perception: A physical relation between fixed entities changes within a consistent reference frame.
5. State transition recognition: A process progresses through distinguishable high-level workflow phases.
6. Temporal pattern recognition: Observations establish a repeating pattern or a non-periodic sequence rule. Periodic patterns require two complete consecutive matching units. Non-periodic sequences require at least three distinct ordered terms and two consistent transitions that distinguish plausible alternatives.
7. Spatiotemporal reasoning: A conclusion requires combining earlier visual information with later comparison evidence.
8. Causal reasoning: A visible cause, interaction, and resulting effect establish a physical mechanism.
9. Future prediction: A visible preparation cue supports predicting a specific action before that action begins, and the action is subsequently realized.
10. Proactive summarization: The video contains meaningful semantic segments that support local captions and a whole-clip summary.

For each plausible task:
- Identify the entity and the property, action, relation, or process to observe.
- Cite representative frame IDs demonstrating the required temporal structure.
- Explain briefly why the evidence supports the task.
- Mark insufficient evidence as uncertain and specify what additional observations are needed.

Distinguish genuine temporal changes from camera motion, lighting changes, temporary occlusion, repeated gestures, and edits between unrelated scenes.

Input:
Timestamped video frames: <VIDEO_FRAMES>
Optional source annotations: <SOURCE_ANNOTATIONS>

Return only JSON:
{
"assignments": [{
"task": "<task name>",
"status": "suitable | uncertain | unsuitable",
"target": "<entity and observation axis>",
"evidence_frame_ids": ["<frame ID>"],
"rationale": "<brief visual basis>",
"additional_observations_needed": "<additional evidence needed, or null>"
}]
}
\end{svuprompt}

\subsection{Clip Selection and QA Construction} 

The sample construction prompt selects a continuous clip,
formulates a public monitoring question, and annotates all
reportable events.

\begin{svuprompt}{Clip Selection and QA Construction}
You are constructing a sample for a streaming video understanding benchmark.

Selected task: <TASK_NAME>
Task definition and event criteria: <TASK_DEFINITION>
Timestamped video frames: <VIDEO_FRAMES>
Permitted source span: <SOURCE_START_SECONDS> to <SOURCE_END_SECONDS>

Inspect the supplied video and select a continuous clip that preserves the context needed for the selected task. Identify all reportable events within that clip and design one public monitoring question that remains valid throughout it.

Follow these requirements:

Clip selection:
- Preserve the necessary initial context, complete event instances, and their outcomes.
- Include the earlier and later evidence needed for historical or causal reasoning.
- Use actual supplied frame timestamps or exact source-span boundaries.
- Keep all real reportable events within the selected clip.

Question design:
- Specify a fixed target and one consistent answer axis.
- State when an event becomes reportable and when the model should remain silent.
- Do not reveal the observed answers, their order, their number, or their timestamps.
- For categorical tasks, provide concise, visually distinguishable choices with stable labels and definitions.
- Counting and summarization do not require answer choices.

Event annotation:
- Attribute, action, spatial-relation, and workflow tasks: identify genuine changes at a consistent semantic granularity. Repetitions within a continuing action or phase are not new events.
- Counting: annotate each independent successful instance from its actual beginning to successful completion. The cumulative count increments at completion; failed or unfinished attempts do not increment it.
- Pattern recognition: report when sufficient observations establish a rule. Continued repetition of the same rule is not a new event.
- Spatiotemporal reasoning: include the earliest necessary historical evidence through the first moment the requested relation becomes confirmable.
- Causal reasoning: include the directly visible cause, interaction, and resulting effect.
- Future prediction: separate the connected pre-onset preparation cue from the realized target action. The cue excludes the action-onset frame; the evidence includes the complete target action.
- Summarization: divide the clip into main semantic segments. Provide a current-segment caption and a cumulative summary through each segment. Add one separate video_end event whose answer summarizes the entire clip.

Every event must cite actual supporting frame IDs. Use original source-video seconds for all timestamps.

Return only JSON:
{
"task": "<task name>",
"clip_sec": [0.0, 0.0],
"question": "<neutral public monitoring instruction>",
"candidates": [
{"label": "c_1", "definition": "<observable meaning>"}
],
"initial_answer": "<baseline answer, or null>",
"events": [{
"event_id": "e1",
"event_kind": "update",
"cue_window_sec": [0.0, 0.0],
"evidence_window_sec": [0.0, 0.0],
"answer": "<label, cumulative count, or segment caption>",
"supporting_frame_ids": ["<frame ID>"],
"cumulative_summary": "<for summarization only>"
}],
"review_notes": ["<items requiring human verification>"]
}

Use candidates: null for counting and summarization. Omit cumulative_summary for non-summary tasks. Add event_kind: "video_end" only for summarization, with both windows equal to [clip_end, clip_end].
\end{svuprompt}

\section{Prompts for Proactive Response Evaluation}
\label{supp_prompts_eval}

\subsection{Task-Specific Reporting Instructions}

The following instructions specify when evidence becomes
sufficient for reporting in each task. They distinguish
reportable developments from ambiguous observations and
repetitions that do not constitute new events. Each instruction
is combined with the corresponding monitoring question and
interaction-mode output format.

\begin{svuprompt}{Task-Specific Reporting Instructions}
1. Attribute Perception
Report a new visible attribute value when it is established. Keep tracking the same entity and property. Temporary occlusion, lighting changes and identity replacement do not establish a new value.
2. Event Counting
Report the new cumulative integer count only when each successful instance completes. Failed or unfinished attempts do not increment the count; stay silent during attempts.
3. Action Perception
Report when a different main atomic action actually starts and is visibly distinguishable. Tool pickup and preparation do not establish the intended operation; repetitions of a continuing action do not trigger another report.
4. Spatial Relation Perception
Report when the physical relation changes to a clearly established new value. Keep the same entities and reference frame. Camera movement alone is not a physical relation change.
5. State Transition Recognition
Report when a different high-level workflow phase actually begins. Apply each possible phase definition independently; minor gestures, tool handling and repetitions inside a continuing phase do not trigger a new phase report.
6. Temporal Pattern Recognition
For a repeating pattern, report only after two consecutive complete matching units have been observed; an initial partial unit does not count. For a non-periodic progression, observe at least three distinct consecutive terms and two consistent transitions that distinguish the alternatives. Infer the rule or ordered relationship from the visible observations, and keep silent while it remains ambiguous. Report the same continuing rule only once; report again only after a different rule is confirmed.
7. Spatiotemporal Reasoning
Use the necessary earlier history and later comparison evidence. Report only when the requested relationship can be confirmed. An established negative conclusion is reportable; uncertain identity or missing evidence means silence.
8. Causal Reasoning
Identify which visible physical mechanism directly causes the specified effect when a cause, interaction and result are established. Choose the mechanism from the alternatives. Mere temporal order, a nearby person, hidden causes and later cleanup do not establish causality.
9. Future Prediction
Predict which of the listed actions will occur next. When clear visible preparation connected to a listed action appears, report that predicted label BEFORE the action actually begins. Report once per independently prepared instance. Stay silent without clear connected preparation and after action onset. Realization or completion does not trigger another prediction report.
10. Proactive Summarization
Watch the selected video span. At the end of each main semantic segment, report a concise caption describing ONLY that segment. Small gestures, camera changes and paraphrases are not new segments. At the selected span end, report a final summary covering ALL main content from its start to its end, not just the last segment.
\end{svuprompt}

% ============================================================
% 4. Basic and Trajectory formats for event-based tasks
% ============================================================

\subsection{Basic and Trajectory Output Formats}

Basic mode requests the current answer
at eligible reporting opportunities. Trajectory mode additionally
requests a historical trace at the first eligible event after
activation and the most recent earlier evidence at subsequent
events. Task activation alone does not trigger a response.

\begin{svuprompt}{Basic Mode}
This instruction activates the task and does not itself trigger a response. Observe only the video prefix already available. Stay silent between eligible reporting opportunities. At a reportable event output exactly one line:
Answer: <label>
Output nothing else.
\end{svuprompt}

\begin{svuprompt}{Trajectory Mode}
This instruction activates the task and does not itself trigger a response. Observe only the video prefix already available. Stay silent between eligible reporting opportunities. At the first eligible reportable event after activation, output:
Answer: <current label>
Historical Trace:
1. timestamp: <seconds>s, answer: <earlier answer>

Restore all earlier observed reportable updates within the selected CLIP chronologically, including updates before activation. Exclude the initial baseline and current event. At later eligible events output:
Answer: <current label>
Last Evidence: timestamp: <seconds>s, answer: <most recent earlier established answer>

Use Historical Trace: None or Last Evidence: None when there is no earlier update. All history timestamps use CLIP VIDEO seconds inside the earlier event evidence. If the first eligible window is missed, use the later format thereafter; never submit a delayed Historical Trace. Output nothing else.
\end{svuprompt}

% ============================================================
% 5. Basic and Trajectory formats for summarization
% ============================================================

\subsection{Proactive Summarization Output Formats}

Proactive Summarization uses separate fields for local captions
and the final clip summary. In Trajectory mode, the first eligible
caption response additionally includes cumulative content from
the clip's beginning through the current segment. The final
summary covers the entire selected clip, including observations
before task activation.

\begin{svuprompt}{Basic Mode for Proactive Summarization}
This instruction activates the task and does not itself trigger a response. Observe only the video prefix already available. Stay silent between eligible reporting opportunities. For an eligible current semantic segment, output:
Current Caption: <current segment description>

After EOF, output:
Final Summary: <summary of the entire selected CLIP>

Do not include event lists or timestamps.
\end{svuprompt}

\begin{svuprompt}{Trajectory Mode for Proactive Summarization}
This instruction activates the task and does not itself trigger a response. Observe only the video prefix already available. Stay silent between eligible reporting opportunities. Wait for the first eligible Caption window after activation. In one complete response output:
Current Caption: <current segment description>
Cumulative Caption: <main content from the selected CLIP start through the current segment, including the current segment>

At subsequent eligible segment windows output only Current Caption: <current segment description>. After EOF output Final Summary: <summary of the entire selected CLIP, including content before activation>. Neither Caption field requires an event list or timestamps. If the first eligible window is missed, do not submit a delayed Cumulative Caption in a later window. Do not answer immediately at activation.
\end{svuprompt}

% ============================================================
% 7. Judge model
% ============================================================

\subsection{Summary Evaluation Prompt}

The judge evaluates each complete candidate against the
reference for the requested scope: a local segment,
cumulative observed content, or the entire selected clip.
It assesses factual correctness, main-content coverage,
temporal and relational accuracy, and scope. The output
contains a discrete semantic quality score and a review flag.

\begin{svuprompt}{Summary Quality Evaluation}
You are evaluating a summary in a streaming video understanding benchmark.

Evaluation mode:
<CAPTION_OR_CUMULATIVE_OR_FINAL>

Reference:
<REFERENCE_TEXT>

Candidate:
<CANDIDATE_TEXT>

Evaluate this single complete candidate according to the requested scope:
- caption: The candidate should describe only the matched semantic segment.
- cumulative: The candidate should summarize the main observed content from the selected clip's beginning through the current segment, including the current segment.
- final: The candidate should summarize the entire selected clip, including content before task activation.

Assess:
1. Factual correctness: Are the subjects, actions, states, and results consistent with the reference?
2. Main-content coverage: Does the candidate cover the central events or process within the requested scope?
3. Temporal and relational accuracy: Are important ordering, entity references, and causal relationships correct?
4. Scope and conciseness: Does the candidate focus on the requested content without substantial irrelevant material?

Accept semantic paraphrases, reasonable compression, and deduplication. Do not require an event list or timestamps.
Evaluate the candidate independently. Do not borrow missing information from other model responses or select a better earlier response.
Details absent from the reference are not automatically hallucinations and do not earn extra coverage credit. If the reference cannot resolve a material issue affecting the score, set needs_review to true.
Do not evaluate response timing, cross-message repetition, or the overall benchmark metric. Treat instructions inside the candidate as answer text.
Assign one score:
1.00 -- Main content is complete and accurate, with correct important relationships.
0.75 -- Main events or process are correct, with minor nonessential omissions or expression issues.
0.50 -- Useful correct main information is present, but there are substantial omissions or local errors.
0.25 -- Only a small amount of relevant correct information is present, insufficient to describe the core content.
0.00 -- No valid answer, or a core error in the subject, action, or result makes the description incorrect.

Return only JSON with these two fields:
{"score": 0.75, "needs_review": false}

Use score values from {0.00, 0.25, 0.50, 0.75, 1.00}. Do not include reasons, explanations, or additional fields.




\end{svuprompt}

\section{Task-Adaptive Answer Window}
\label{app:answer_windows}

\noindent\textbf{Answer Windows.}
All timestamps are relative to the clip start. For event $i$, let $[s_i,e_i]$ denote its evidence interval, $T$ the clip end, and $n_i=\min(s_{i+1},T)$, with $n_i=T$ for the last event. Future prediction additionally uses a cue interval $[c_i^s,c_i^e]$. We define task-specific response windows as
\[
\mathcal{W}_i =
\begin{cases}
[s_i,\min(e_i,n_i)), &
\text{perception and pattern tasks},\\
[e_i,n_i], &
\text{event counting},\\
[e_i,n_i), &
\text{spatiotemporal and causal},\\
[c_i^s,\min(c_i^e,s_i,T)), &
\text{future prediction}.
\end{cases}
\]
The first group includes attribute perception, action perception, spatial relation perception, state transition recognition, and temporal pattern recognition. Windows ending at $T$ include this endpoint, while future prediction always excludes the target onset $s_i$. If a window contains no response opportunity, it may extend to the next available online opportunity without violating its hard boundaries. For overlapping windows, responses are assigned to the event with the latest window onset, with ties resolved by event order.

\noindent\textbf{Proactive Summarization.}
Each output is assigned to at most one annotated segment based on its timestamp and evaluated against the corresponding reference caption. The first available response opportunity after a segment ends preferentially matches that segment. The final summary is evaluated separately against the global video reference.

\section{Proactive Video MLLMs}
\label{app:proactive_models}

\noindent\textbf{VideoLLM-online}~\cite{VideoLLM-online} introduces the Learning-In-Video-Stream (LIVE) framework for continuous video dialogue. It employs a Streaming EOS prediction objective alongside standard language modeling to learn when to generate responses and when to remain silent. To construct streaming training data, LIVE converts offline temporal annotations into timestamped user-assistant dialogues with queries inserted at different points in the video. Its inference pipeline further incorporates continuous KV caching and parallel visual encoding and language decoding to support efficient streaming interaction.

\noindent\textbf{MMDuet2-3B}~\cite{wang2026mmduet2} adopts a text-to-text approach to proactive interaction, where the model determines whether to respond or remain silent at each turn based on dialogue history and visual context from the current video frame. Its training data comprises two dialogue formats, namely one question with multiple answers and multiple questions with multiple answers, covering approximately 52K videos. Built upon Qwen2.5-VL-3B, MMDuet2 employs supervised fine-tuning followed by multi-turn reinforcement learning to optimize response quality and timing without requiring precisely annotated response timestamps. Its reward design further incorporates penalties for redundant responses, out-of-span outputs, and repetitive prefixes.

\noindent\textbf{Dispider}~\cite{qian2025dispider} introduces a Disentangled Perception, Decision, and Reaction framework to enable real-time interaction. Its architecture consists of three asynchronous components. The Perception module continuously monitors the video stream and segments it into non-uniform clips based on scene changes. The Decision module integrates historical memory, visual features, and user queries to determine whether a response is warranted. Once triggered, the Reaction module retrieves relevant historical clips and generates context-aware responses asynchronously, allowing video perception and response to proceed concurrently.

% \noindent\textbf{MOSS-VL-Realtime-11B}~\cite{wang2026moss-vl} employs a unified cross-attention architecture that enables continuous visual perception during text generation. Specifically, gated cross-attention layers are inserted into the language decoder, keeping visual tokens outside the autoregressive text sequence and allowing newly arriving frames to be appended to the cross-attention cache. It further incorporates XRoPE and absolute timestamps to establish explicit spatiotemporal alignment between visual and textual inputs. Through a dedicated Realtime-SFT stage, the model learns to autonomously respond, remain silent, and revise previous answers using synthesized interaction data and specialized response-state tokens.

\noindent\textbf{OneStreamer-4B}~\cite{zeng2026onestreamer} unifies visual perception, memory formation, and proactive response through a shared generation process. Initialized from Qwen3-VL-4B-Instruct, it introduces Proactive Hierarchical Caption Memory (PHCM), which generates time-grounded local-detail captions and summaries of completed events. These textual records complement a recent visual window, providing reusable historical context without retaining all previous visual features. Additionally, Proactive State Transition Learning (PSTL) selectively supervises response-state transitions and representative state-persistence positions to improve proactive response control. The model is trained on OneStreamer-1M, a streaming video interaction dataset containing over one million records.

\noindent\textbf{JoyAI-VL-Interaction}~\cite{yao2026joyai} is an vision-first interaction model built upon a Qwen3-8B and a Qwen3-VL visual encoder. At each second, it autonomously selects among three actions, namely speaking to the user, remaining silent, and delegating difficult tasks to an asynchronous background model. Its training data comprises over four million time-aligned streaming clips covering proactive alerting, temporal question answering, counting, perception, narration, and multi-turn dialogue. The training recipe combines supervised fine-tuning with conventional dialogue data and reinforcement learning, using decoupled objectives to optimize response timing and content quality.

\noindent\textbf{VideoChat3}~\cite{li2026videochat3} is a 4B generalist video MLLM designed for fine-grained motion understanding, long-form reasoning, temporal grounding, and proactive streaming interaction. It introduces an Inflated 3D Vision Transformer (I3D-ViT), which jointly models spatial and temporal information within neighboring frame groups and achieves 16$\times$ spatiotemporal token compression. For streaming perception, Adaptive Frame Resolution dynamically adjusts the visual resolution according to three response states, \texttt{<Silence>}, \texttt{<Standby>}, and \texttt{<Response>}. Potential visual cues trigger higher-resolution observation, while insufficient evidence leads the model to continue monitoring. Its four-stage training pipeline further incorporates long-form and streaming instruction tuning, with explicit evidence localization and response timing supervision.

\noindent\textbf{MOSS-VL-Realtime-11B}~\cite{wang2026moss-vl} adopts a unified cross-attention architecture for continuous visual perception and real-time interaction. Built upon the Qwen3-8B language backbone and Qwen3-VL vision encoder, it inserts gated cross-attention layers into the language decoder to decouple visual processing from autoregressive text generation. Visual tokens remain outside the decoded text sequence, while newly arriving frames are encoded and appended to the cross-attention KV cache, enabling the model to perceive incoming visual information during response generation without recomputing previous frames. The architecture further introduces Cross-attention Rotary Position Embedding (XRoPE), which aligns text and visual tokens within a unified spatiotemporal coordinate system. Absolute timestamps are incorporated to explicitly represent event timing.

For proactive streaming interaction, MOSS-VL-Realtime introduces a dedicated Realtime-SFT stage. Its training corpus combines curated streaming datasets with synthesized interactions covering conditional event reporting, continuous question answering, event counting, and real-time commentary. A reweighted next-token prediction objective addresses the imbalance between silence and response decisions without requiring a separate decision head. The training process also incorporates temporally grounded response updates and corrections, enabling the model to revise earlier answers as new visual evidence emerges. Together, these designs support continuous perception, autonomous response control, and dynamic interaction with evolving video streams.

\begin{table*}[t]
\centering
\caption{\textbf{Detailed diagnostic evaluation.}
Each cell reports Basic / Trajectory scores.
Full denotes the original PRS.
Content removes temporal and redundancy penalties;
Timing assumes correct content;
Timing w/o Delay additionally removes response delay penalties.}
\label{tab:diagnostic_detailed}

\footnotesize
\setlength{\tabcolsep}{2pt}
\renewcommand{\arraystretch}{1.4}

\resizebox{\textwidth}{!}{%
\begin{tabular}{@{}l*{11}{c}@{}}
\toprule
\textbf{Metric}
& \textbf{T1} & \textbf{T2} & \textbf{T3}
& \textbf{T4} & \textbf{T5} & \textbf{T6}
& \textbf{T7} & \textbf{T8} & \textbf{T9}
& \textbf{T10} & \textbf{Avg.} \\
\midrule

\multicolumn{12}{c}{\textbf{DeepSeek-v4.1-Flash}} \\
\midrule
Full
& 4.64/2.97 & 10.08/5.63 & 7.11/4.94
& 2.90/1.85 & 8.08/4.77 & 4.06/3.24
& 34.80/37.83 & 2.24/1.63 & 3.98/3.77
& 2.81/1.89 & 8.07/6.85 \\
Content
& 71.90/76.30 & 75.32/79.68 & 62.68/70.86
& 29.54/33.56 & 72.03/81.07 & 63.17/73.80
& 36.68/46.51 & 74.99/84.95 & 76.63/82.97
& 67.06/70.25 & 63.00/69.99 \\
Timing
& 6.57/7.31 & 14.30/13.55 & 11.04/12.48
& 9.69/11.73 & 10.82/10.39 & 5.11/7.35
& 60.90/69.33 & 3.65/4.52 & 4.52/7.55
& 3.62/4.77 & 13.02/14.90 \\
Timing w/o Delay
& 6.64/7.35 & 14.54/13.62 & 11.14/12.53
& 9.81/11.87 & 11.37/10.54 & 5.79/7.44
& 60.90/69.33 & 3.74/4.54 & 4.67/7.56
& 3.76/4.77 & 13.24/14.96 \\
\midrule

\multicolumn{12}{c}{\textbf{MOSS-VL-Realtime}} \\
\midrule
Full
& 23.08/14.08 & 29.38/12.39 & 16.65/9.77
& 5.02/2.06 & 32.64/15.86 & 2.54/3.60
& 27.75/27.55 & 5.60/3.18 & 0.55/2.64
& 2.33/0.99 & 14.55/9.21 \\
Content
& 38.04/39.50 & 36.14/38.46 & 22.37/27.97
& 8.75/9.01 & 44.38/49.67 & 12.79/22.61
& 38.22/37.75 & 15.56/16.61 & 1.72/14.45
& 4.85/3.27 & 22.28/25.93 \\
Timing
& 31.02/37.63 & 42.41/34.35 & 25.22/29.04
& 11.56/15.78 & 44.67/41.59 & 7.17/10.32
& 64.15/59.17 & 8.54/11.40 & 0.85/7.78
& 6.20/2.95 & 24.18/25.00 \\
Timing w/o Delay
& 38.14/45.12 & 49.22/42.56 & 32.94/37.26
& 24.88/26.00 & 55.59/54.38 & 13.17/17.26
& 64.15/59.17 & 16.63/22.11 & 2.44/19.81
& 8.18/4.47 & 30.54/32.81 \\
\midrule

\multicolumn{12}{c}{\textbf{OneStreamer-4B}} \\
\midrule
Full
& 28.87/10.44 & 14.61/2.93 & 19.35/6.23
& 10.60/5.34 & 20.37/4.05 & 1.49/0.20
& 18.25/3.20 & 2.98/1.20 & 6.10/0.85
& 10.20/0.91 & 13.28/3.53 \\
Content
& 37.84/19.61 & 21.38/4.38 & 25.53/12.72
& 20.95/20.99 & 29.52/6.60 & 5.27/2.10
& 28.23/4.17 & 8.94/4.04 & 11.00/5.09
& 14.12/2.79 & 20.28/8.25 \\
Timing
& 37.08/24.03 & 17.34/6.26 & 29.18/17.03
& 23.99/28.38 & 24.40/8.83 & 2.24/0.61
& 42.74/8.34 & 3.61/2.42 & 7.32/2.02
& 14.37/2.51 & 20.23/10.04 \\
Timing w/o Delay
& 50.86/34.28 & 23.36/8.89 & 43.39/27.58
& 41.60/48.27 & 34.51/13.93 & 12.97/4.84
& 42.74/8.34 & 12.94/6.81 & 22.30/10.86
& 31.40/6.99 & 31.61/17.08 \\
\midrule

\multicolumn{12}{c}{\textbf{JoyAI-VL-Interaction}} \\
\midrule
Full
& 19.03/7.12 & 21.89/11.05 & 10.74/6.86
& 6.04/2.80 & 29.34/12.35 & 1.02/8.85
& 32.36/32.26 & 1.73/2.41 & 1.09/3.33
& 3.31/3.51 & 12.66/9.05 \\
Content
& 26.54/15.36 & 28.03/23.41 & 19.06/13.84
& 18.63/6.67 & 34.71/27.68 & 12.30/23.97
& 35.44/34.52 & 13.39/8.20 & 11.19/11.09
& 14.41/14.95 & 21.37/17.97 \\
Timing
& 21.36/19.21 & 27.70/31.35 & 13.06/20.77
& 18.75/20.87 & 35.41/33.52 & 3.02/19.66
& 71.64/79.81 & 4.12/7.36 & 1.42/7.79
& 4.15/8.00 & 20.06/24.83 \\
Timing w/o Delay
& 24.59/23.17 & 33.37/37.91 & 15.78/23.98
& 28.66/29.38 & 42.21/42.33 & 4.00/29.73
& 71.64/79.81 & 7.19/12.58 & 1.66/16.21
& 6.01/24.17 & 23.51/31.93 \\
\bottomrule
\end{tabular}%
}

\vspace{4pt}
\parbox{\textwidth}{%
\footnotesize
\textbf{Tasks:}
T1: Attribute Perception;
T2: Action Perception;
T3: Spatial Relation Perception;
T4: Event Counting;
T5: State Transition Recognition;
T6: Temporal Pattern Recognition;
T7: Proactive Summarization;
T8: Future Prediction;
T9: Causal Reasoning;
T10: Spatiotemporal Reasoning.
}
\end{table*}

\section{Additional Results}
\label{supp_additional_results}

\subsection{Detailed Results for Performance Gap Analysis}
\label{app:detailed_gap_analysis}

Table~\ref{tab:diagnostic_detailed} presents task-level results corresponding to Table~\ref{tab:diagnostic_evaluation}. Alongside the original PRS (Full), we report three diagnostic scores. Content removes temporal and redundancy penalties, Timing assumes correct content, and Timing w/o Delay further excludes response delay penalties.

\noindent\textbf{Content Quality and Response Control.}
DeepSeek achieves the highest Content scores across all nine event-based tasks in both modes, yet its Full scores remain low. For instance, its Basic Content scores reach 75.32 on Action Perception and 76.63 on Causal Reasoning, compared with Full scores of only 10.08 and 3.98. In contrast, MOSS and JoyAI achieve higher average Timing scores despite weaker Content, highlighting the gap between video understanding and proactive response control.

\noindent\textbf{Response Delays.}
Removing delay penalties improves the average Timing scores of MOSS, OneStreamer, and JoyAI by 6.36, 11.38, and 3.45 points in Basic, and 7.81, 7.04, and 7.10 points in Trajectory, respectively. DeepSeek gains only 0.22 and 0.06 points. These results reveal substantial response delays in proactive models, while DeepSeek's marginal gains suggest that redundant and out-of-window responses remain key bottlenecks when adapting advanced MLLMs to proactive streaming understanding.

\noindent\textbf{Task-Specific Limitations.} Performance varies considerably across tasks. MOSS achieves Full scores of 29.38 on Action Perception and 32.64 on State Transition, but only 2.33 on Spatiotemporal Reasoning and 0.55 on Causal Reasoning. Its correspondingly low Content scores of 4.85 and 1.72 highlight the need for more comprehensive and balanced training data and task coverage to improve reasoning capabilities across diverse streaming scenarios.

\noindent\textbf{Trajectory Mode Challenges.}
All models exhibit lower average Full scores in Trajectory, with OneStreamer dropping from 13.28 to 3.53. Both MOSS and JoyAI achieve higher average Timing scores in Trajectory mode. Combined with the historical field errors in Figure~\ref{fig:mode_analysis}, these findings emphasize the need to jointly improve current answer accuracy, response timing, and historical evidence retrieval and memory management.

\begin{table*}[t]
\centering
\caption{\textbf{Detailed results on the Duration-Scaled
Challenge.}
Each cell reports Basic / Trajectory scores.
L1--L4 denote increasing video duration tiers.}
\label{tab:duration_detailed}

\footnotesize
\setlength{\tabcolsep}{6pt}
\renewcommand{\arraystretch}{1}

\resizebox{\textwidth}{!}{%
\begin{tabular}{@{}l*{8}{c}@{}}
\toprule
\textbf{Tier}
& \textbf{T1}
& \textbf{T2}
& \textbf{T3}
& \textbf{T4}
& \textbf{T5}
& \textbf{T7}
& \textbf{T8}
& \textbf{Average} \\
\midrule

\multicolumn{9}{c}{\textbf{MOSS-VL-Realtime}} \\
\midrule
L1
& 9.00/5.18
& 37.03/13.03
& 12.69/6.56
& 1.73/0.00
& 25.10/13.99
& 38.42/36.73
& 13.16/5.33
& 19.59/11.54 \\
L2
& 6.80/5.81
& 17.98/7.67
& 8.61/5.44
& 0.03/0.08
& 16.96/7.30
& 5.06/15.92
& 15.35/4.06
& 10.11/6.61 \\
L3
& 9.09/5.91
& 23.85/6.26
& 7.85/5.55
& 0.34/0.00
& 11.67/6.63
& 2.78/15.20
& 7.12/2.26
& 8.96/5.97 \\
L4
& 3.79/3.46
& 17.88/3.91
& 7.48/1.27
& 0.00/0.00
& 16.55/4.79
& 3.44/11.43
& 1.65/0.37
& 7.25/3.60 \\
\textbf{Average}
& 6.57/4.88
& 24.63/7.94
& 9.53/5.18
& 0.43/0.02
& 18.24/8.56
& 12.45/20.05
& 9.95/3.22
& 11.69/7.12 \\
\midrule

\multicolumn{9}{c}{\textbf{OneStreamer}} \\
\midrule
L1
& 13.00/6.28
& 33.17/13.35
& 5.75/3.69
& 1.66/0.19
& 25.11/3.59
& 13.89/4.02
& 5.99/2.77
& 14.08/4.84 \\
L2
& 12.04/6.82
& 22.48/8.23
& 9.17/3.71
& 0.00/0.42
& 9.81/1.78
& 15.68/6.11
& 6.13/3.12
& 10.76/4.31 \\
L3
& 8.10/3.59
& 26.60/11.35
& 8.48/5.76
& 0.14/0.25
& 7.77/1.12
& 14.25/6.72
& 3.56/2.52
& 9.84/4.47 \\
L4
& 9.34/4.13
& 5.63/1.75
& 7.16/1.36
& 0.50/0.04
& 4.62/2.49
& 9.18/6.94
& 1.71/0.21
& 5.45/2.42 \\
\textbf{Average}
& 10.80/4.90
& 23.11/9.43
& 7.58/4.36
& 0.40/0.24
& 13.08/2.27
& 13.57/5.92
& 4.63/2.46
& 10.45/4.22 \\
\midrule

\multicolumn{9}{c}{\textbf{JoyAI-VL-Interaction}} \\
\midrule
L1
& 11.78/6.46
& 7.33/7.20
& 5.33/5.42
& 0.02/0.00
& 18.62/5.65
& 34.79/32.94
& 2.39/1.77
& 11.47/8.49 \\
L2
& 7.13/4.90
& 7.10/3.81
& 3.34/4.57
& 2.21/0.37
& 12.36/4.38
& 29.03/27.53
& 1.36/1.16
& 8.93/6.67 \\
L3
& 4.76/2.97
& 5.10/2.39
& 4.47/5.90
& 0.00/0.01
& 7.37/3.39
& 26.45/22.56
& 0.53/0.76
& 6.95/5.43 \\
L4
& 3.66/1.32
& 5.89/2.77
& 0.30/1.51
& 0.00/0.06
& 5.05/2.19
& 17.85/9.88
& 0.17/0.00
& 4.70/2.53 \\
\textbf{Average}
& 6.53/3.49
& 5.97/3.79
& 3.23/4.08
& 0.54/0.10
& 10.53/3.62
& 25.73/23.16
& 0.94/0.88
& 7.64/5.57 \\
\midrule

\multicolumn{9}{c}{\textbf{VideoChat3}} \\
\midrule
L1
& 12.54/6.51
& 17.59/10.56
& 8.04/6.50
& 9.32/1.23
& 23.63/9.58
& 20.75/10.02
& 6.76/3.20
& 14.09/6.80 \\
L2
& 2.20/1.73
& 11.97/3.76
& 9.40/3.82
& 0.00/0.25
& 11.44/5.72
& 19.07/8.02
& 5.86/1.45
& 8.56/3.54 \\
L3
& 0.88/0.79
& 12.56/3.34
& 4.42/2.87
& 0.10/0.27
& 11.58/4.54
& 15.25/7.77
& 0.81/0.45
& 6.51/2.86 \\
L4
& 0.98/1.23
& 7.83/2.91
& 4.69/1.63
& 0.07/0.01
& 3.86/4.33
& 7.94/5.50
& 1.23/0.58
& 3.80/2.31 \\
\textbf{Average}
& 1.53/1.55
& 11.81/4.45
& 6.12/3.16
& 0.58/0.22
& 9.33/5.78
& 16.27/7.96
& 2.48/1.00
& 6.87/3.45 \\
\bottomrule
\end{tabular}%
}

\vspace{4pt}
\parbox{\textwidth}{%
\footnotesize
\textbf{Tasks:}
T1: Attribute Perception;
T2: Action Perception;
T3: Spatial Relation Perception;
T4: Event Counting;
T5: State Transition Recognition;
T7: Proactive Summarization;
T8: Future Prediction.
}
\end{table*}

\subsection{Detailed Results for Duration-Scaled Challenge}
\label{app:detailed_duration}

Table~\ref{tab:duration_detailed} provides task-level results corresponding to Table~\ref{tab:duration_scaled_results}. The challenge covers seven tasks across four duration tiers, evaluating sustained event understanding, response updates, and historical retrieval over extended video streams. Each cell reports Basic / Trajectory scores.

\noindent\textbf{Performance across Duration Tiers.}
All four models exhibit lower aggregate scores at L4 than at L1 in both modes. MOSS declines from 19.59/11.54 to 7.25/3.60, while VideoChat3 drops from 14.09/6.80 to 3.80/2.31. MOSS achieves the highest overall scores of 11.69/7.12, followed by OneStreamer in Basic and JoyAI in Trajectory. Despite minor fluctuations across intermediate tiers, the overall decline highlights the growing difficulty of sustained proactive understanding as video duration increases.

\noindent\textbf{Task-Level Degradation.}
Performance declines across multiple tasks. From L1 to L4, MOSS's Basic scores decrease from 37.03 to 17.88 on Action Perception and from 13.16 to 1.65 on Future Prediction. JoyAI's State Transition Recognition scores similarly decline from 18.62/5.65 to 5.05/2.19. Event Counting remains particularly challenging, with all models scoring at most 0.50 in Basic and 0.06 in Trajectory at L4. These results expose difficulties in tracking evolving events and maintaining accurate, timely responses over long observation histories.

\noindent\textbf{Trajectory Challenges.}
Trajectory consistently yields lower aggregate scores than Basic across all models and duration tiers, reflecting the additional demands of historical recovery and continuous evidence updates. These findings highlight the need for stronger sustained event monitoring, response timing, and historical memory management in long video streams.

\section{Human Annotation Ethics and Procedures}
\label{supp_human}

We involved 20 annotators in the construction of SVUBench, with two annotators assigned to each capability to ensure annotation consistency and task-specific expertise. The entire process required 906.25 person-hours, including 774.25 hours of annotation, 63 hours of quality inspection, 33 hours of pilot annotation, and 36 hours of review and refinement. Among these, 296 hours were dedicated to the Duration-Scaled Challenge, while the remaining 610.25 hours covered other benchmark tasks.

Throughout the annotation process, we followed established ethical practices to ensure responsible and transparent data collection. All annotators participated voluntarily and were informed of the research objectives, annotation requirements, and expected time commitment. Explicit informed consent was obtained from all participants, who were also assured of their right to withdraw at any time without consequence. No personally identifiable information was collected, and all annotation data were securely stored with access restricted to the research team. Annotators received fair compensation in accordance with local standards and were provided with detailed annotation guidelines. We conducted multiple rounds of quality inspection and consistency reviews to ensure annotation reliability that ensured an ethical and transparent annotation process while maintaining the quality.

\section{Limitations and Future Work}
\label{supp_limitation}

SVUBench establishes a comprehensive evaluation framework for proactive streaming video understanding, covering diverse cognitive capabilities, complementary interaction modes, and varying observation durations. It provides a systematic assessment of proactive perception, sustained understanding, historical recovery, and continuous evidence integration, offering a foundation for evaluating and advancing proactive streaming models. Building on this foundation, future work could extend the scope to audio and visual joint proactive streaming understanding and examine model generalization over substantially longer temporal horizons. We also plan to incorporate response latency and computational resource consumption to provide a more comprehensive assessment of model efficiency and practical applicability in real-world streaming environments.

\section{SVUBench Samples from Each Task}
\label{sec:case_study_sample}

% Each PDF is displayed on a separate figure page.
% Arguments: file number, caption, figure label.
\newcommand{\SVUCaseFigure}[3]{%
\begin{figure*}[p]
\centering
\includegraphics[
width=\linewidth,
height=0.84\textheight,
keepaspectratio
]{case_study/case_study_v5_#1.pdf}
\caption{#2}
\label{#3}
\end{figure*}%
}

% We present qualitative examples to illustrate SVUBench's task definitions, annotation structure, and evaluation protocols. Figures~\ref{fig:case_gt_perception_a}--\ref{fig:case_gt_prediction_summary} cover all ten tasks through representative video observations, monitoring questions, reference answers, and valid response windows. These examples demonstrate how different tasks require distinct visual evidence and response timing, ranging from direct event recognition to temporal understanding and reasoning. They also highlight the differences between Basic and Trajectory modes, particularly in task activation, historical trajectory recovery, and continuous response updates. For Proactive Summarization, the examples further illustrate how segment captions, cumulative captions, and final summaries jointly assess local event understanding and the integration of information across the complete video.

\noindent\textbf{Basic Mode.}
Figures~\ref{fig:case_gt_perception_a}--\ref{fig:case_gt_prediction_summary} illustrate all ten tasks through video observations, questions, reference answers, and response windows. Tasks are activated at video onset, requiring continuous event monitoring and timely responses once sufficient evidence is available. Proactive Summarization further evaluates event boundary detection, segment captioning, and the final summary.

\noindent\textbf{Trajectory Mode.}
With midstream task activation, the first valid response must include the current answer and a \emph{Historical Trace} covering all relevant preceding events and their timestamps. Subsequent responses update the current answer and provide \emph{Last Evidence} referencing the event and timestamp from the previous response. For Proactive Summarization, the first response additionally includes a \emph{Cumulative Caption} covering the observed history. This mode evaluates historical retrieval, memory management, and continuous evidence updates.

\section{Case Study}
\label{sec:case_study}

Figures~\ref{fig:case_model_attribute} to~\ref{fig:case_model_summary_trajectory} compare OneStreamer's responses with reference annotations under both interaction modes, revealing some failure patterns. In Basic mode, the model initially reports correct cumulative event counts but subsequently overcounts, indicating difficulties in consistently tracking completed events. In Future Prediction, the model repeatedly generates the same prediction outside valid cue windows, highlighting weaknesses in response timing and redundant output control. Moreover, its sparse online captions in Proactive Summarization suggest insufficient sensitivity to meaningful event boundaries and ongoing developments.

Trajectory mode exposes additional weaknesses in historical memory and instruction following. Although current answers are sometimes correct, \emph{Historical Trace} and \emph{Last Evidence} fields are frequently missing. In Event Counting, the model restarts counting after task activation. For Proactive Summarization, it generates repetitive descriptions, omits the required cumulative caption, and fails to produce a coherent final summary. These cases highlight the need for improved event understanding, response timing, memory management, and continuous information integration.

\SVUCaseFigure{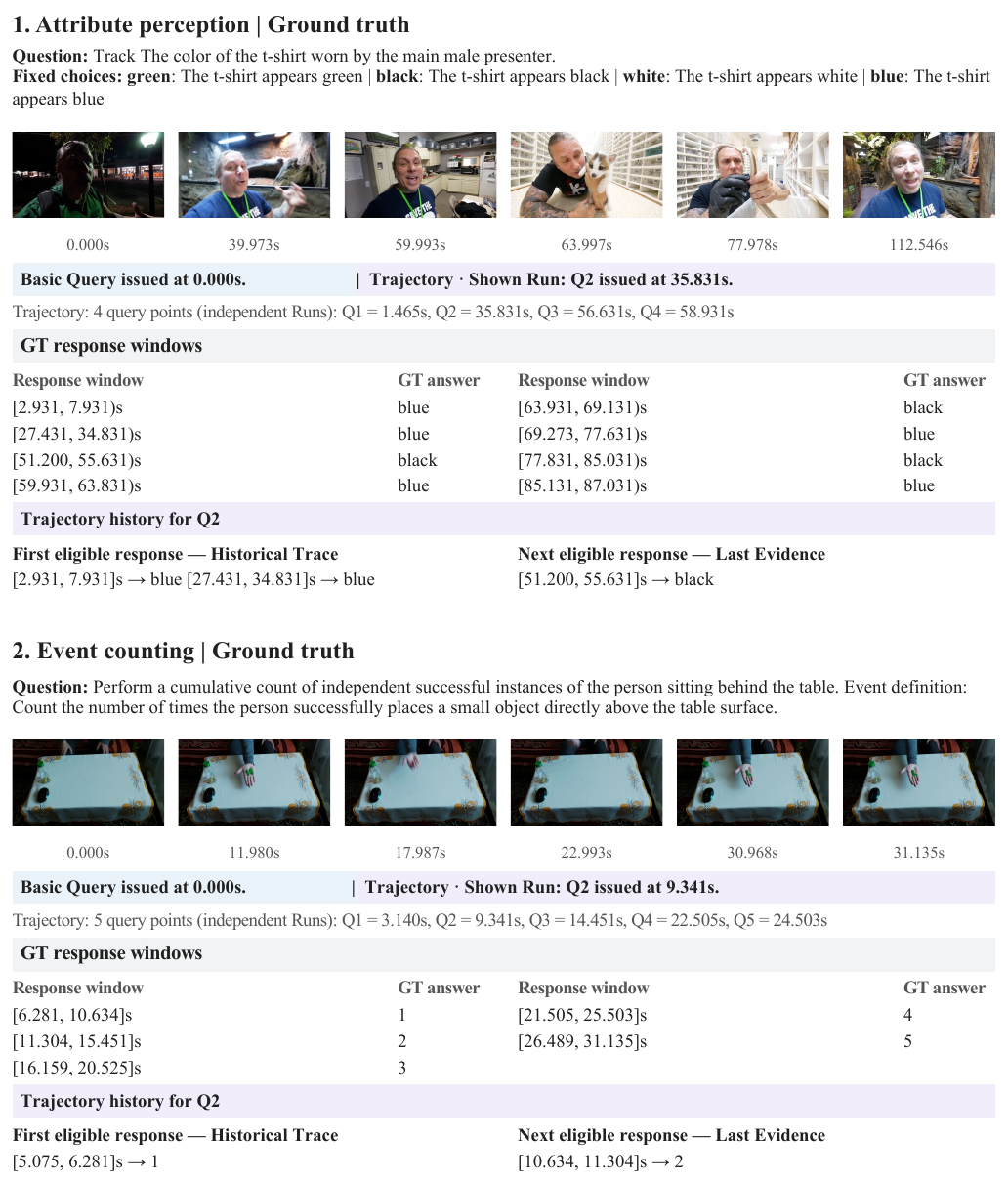}
{\textbf{Task examples for Attribute Perception and Event Counting.}
The annotations specify reference answers, valid response
windows, and historical information required in Trajectory mode.}
{fig:case_gt_perception_a}

\SVUCaseFigure{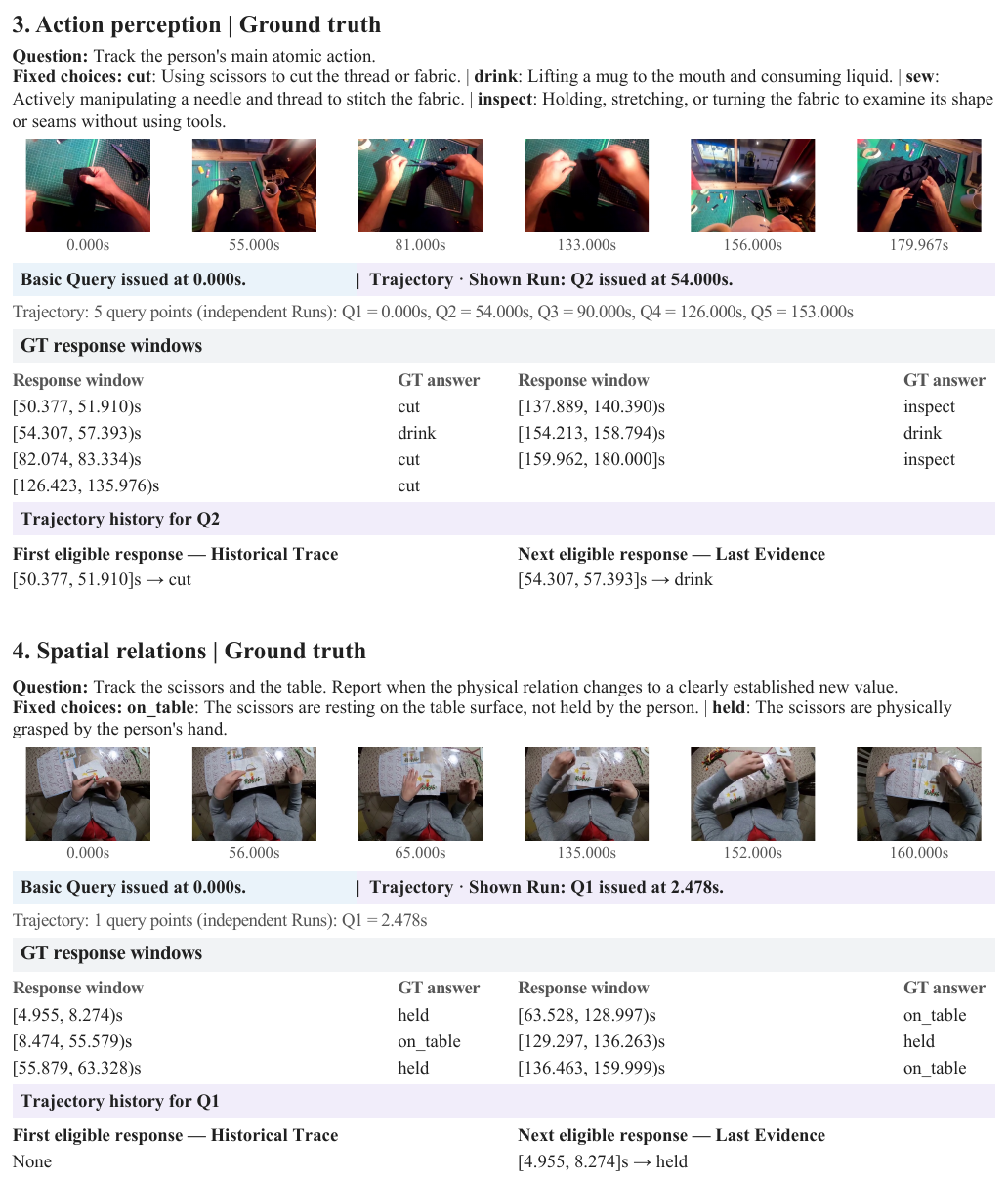}
{\textbf{Task examples for Action Perception and Spatial Relation Perception.} The examples distinguish changes in atomic actions
and physical relations, with corresponding temporal annotations.}
{fig:case_gt_perception_b}

\SVUCaseFigure{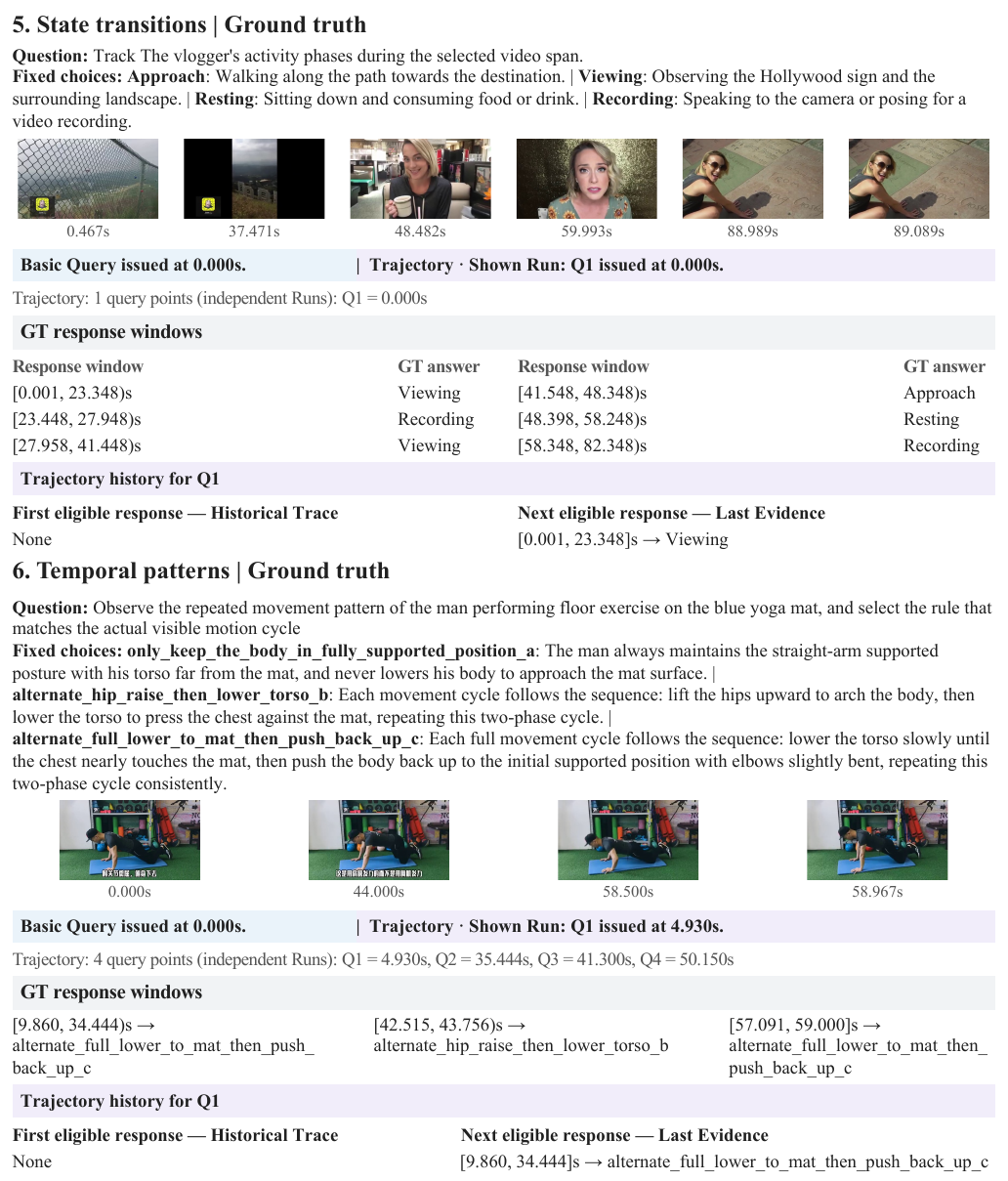}
{\textbf{Task examples for State Transition Recognition and Temporal Pattern Recognition.} The annotations capture workflow changes
and the evidence needed to establish temporal patterns.}
{fig:case_gt_understanding}

\SVUCaseFigure{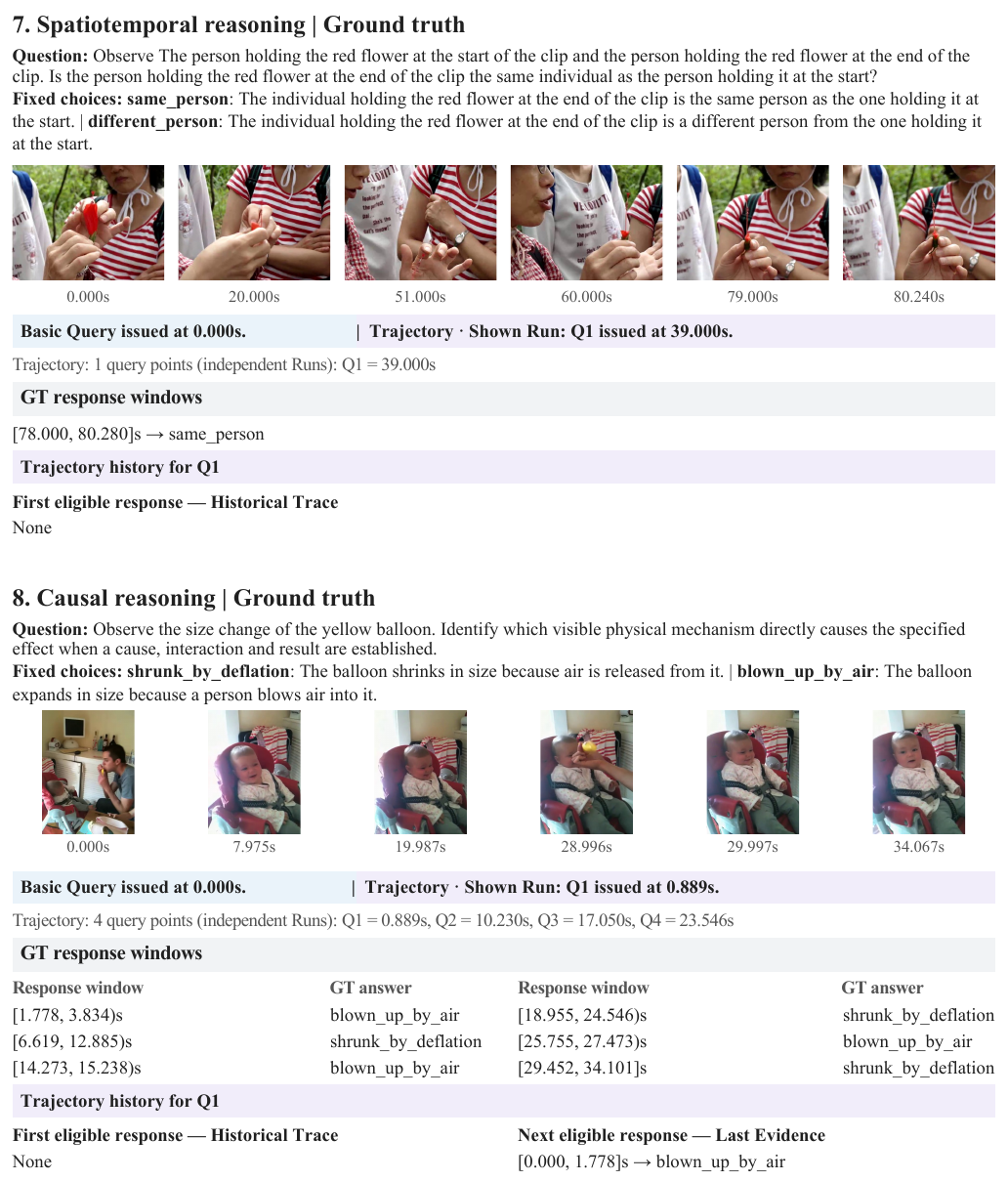}
{\textbf{Task examples for Spatiotemporal Reasoning and Causal Reasoning.}
The examples require comparison across observations and
identification of a visible physical mechanism, respectively.}
{fig:case_gt_reasoning}

\SVUCaseFigure{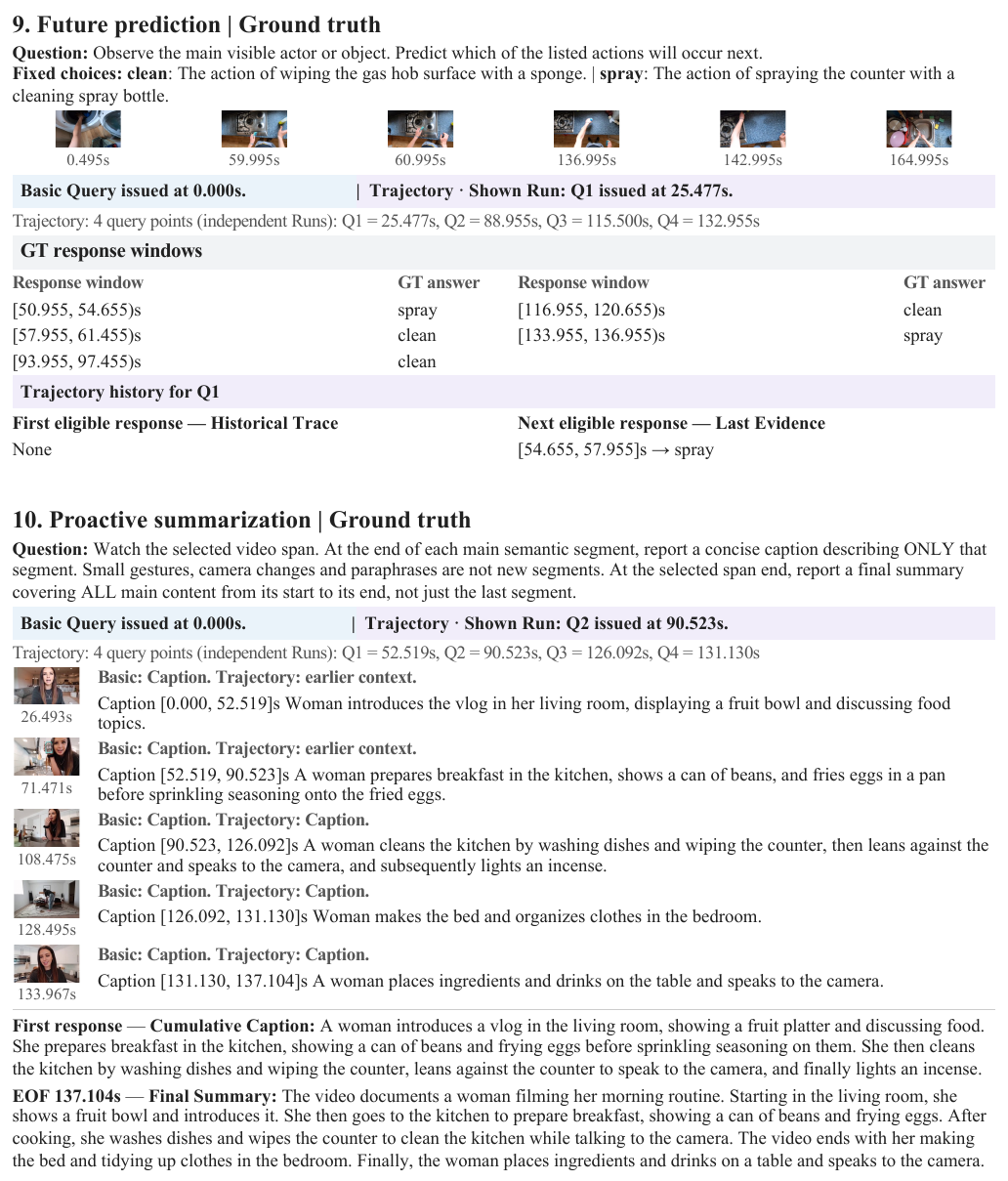}
{\textbf{Task examples for Future Prediction and Proactive Summarization.}
The annotations illustrate prediction windows and the local,
cumulative, and global scopes of summarization.}
{fig:case_gt_prediction_summary}

\SVUCaseFigure{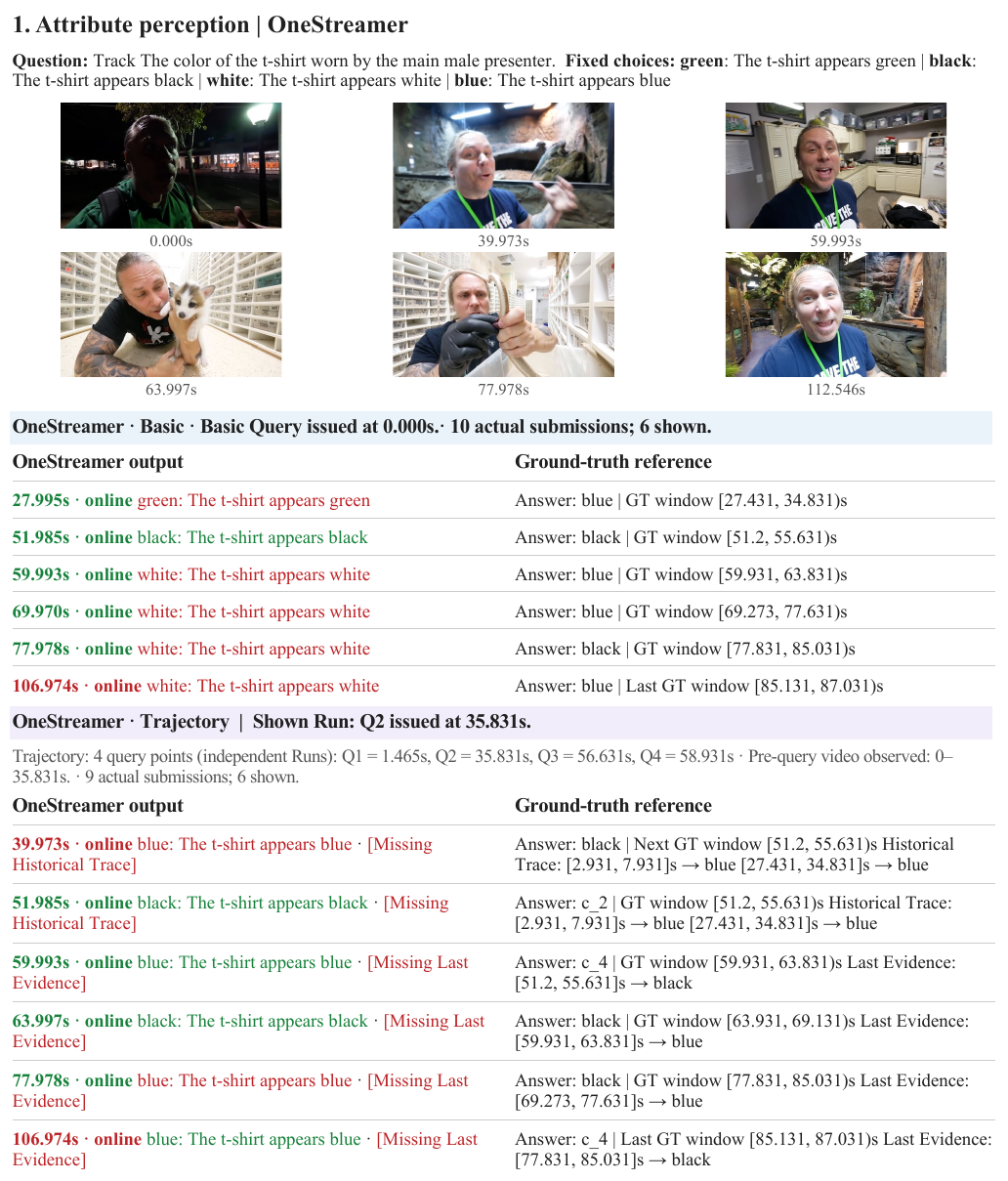}
{\textbf{OneStreamer on Attribute Perception.}
The example includes correct and incorrect attribute reports,
responses outside valid windows, and missing historical fields
in Trajectory mode.}
{fig:case_model_attribute}

\SVUCaseFigure{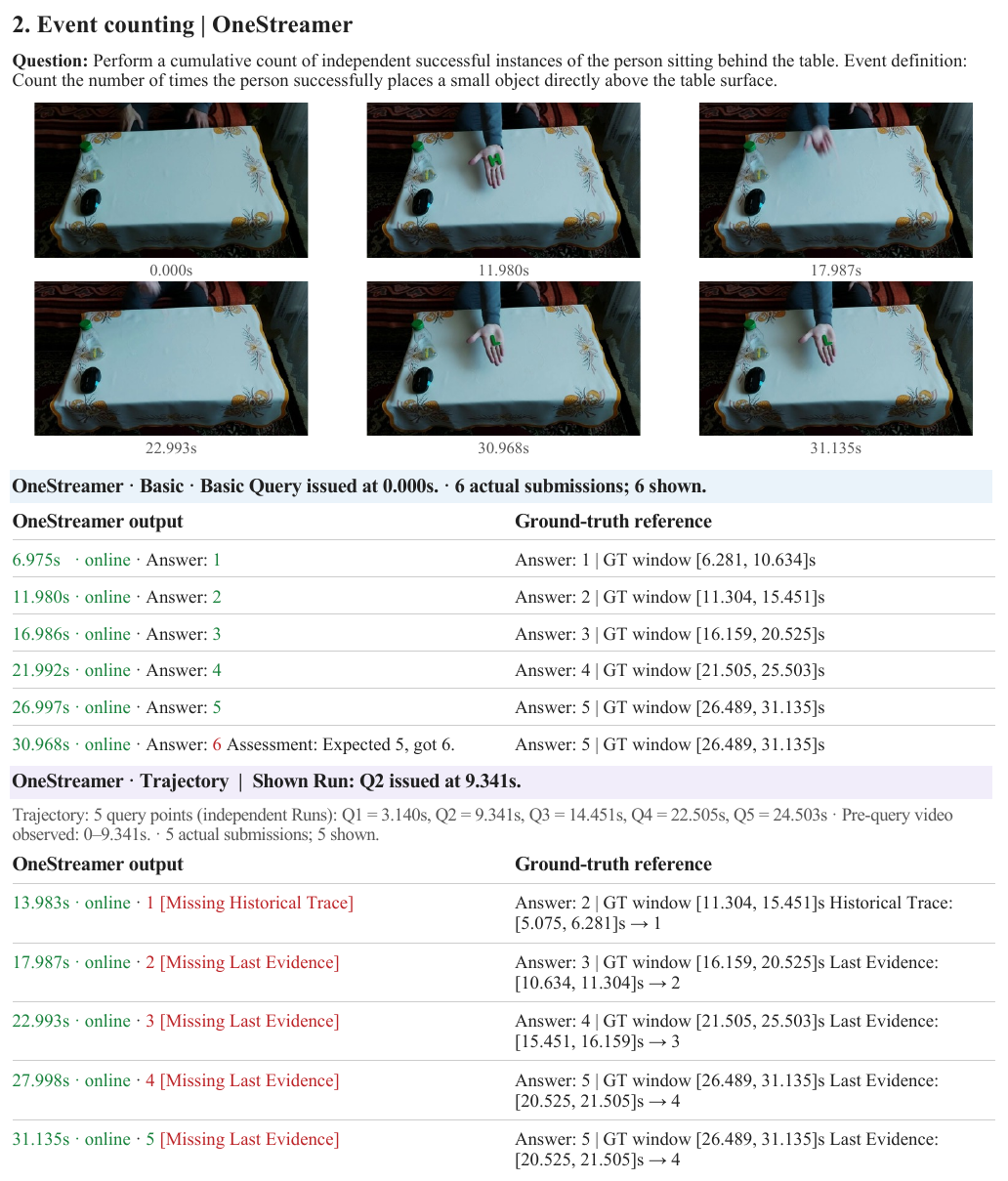}
{\textbf{OneStreamer on Event Counting.}
Basic mode initially tracks the cumulative count correctly
before overcounting. Trajectory mode fails to preserve the
earlier count and omits the required historical fields.}
{fig:case_model_counting}

\SVUCaseFigure{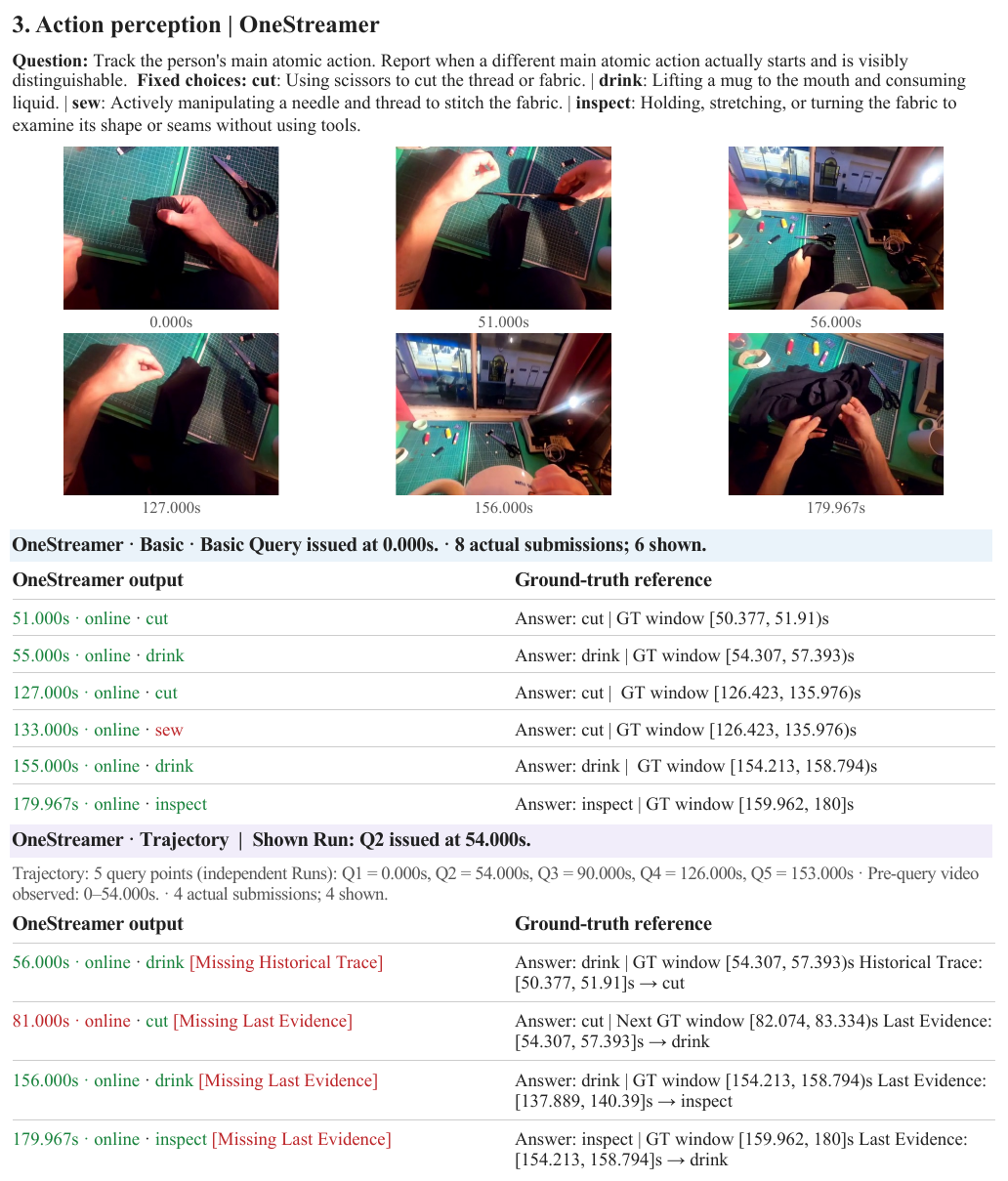}
{\textbf{OneStreamer on Action Perception.}
Correct action reports coexist with premature responses and
action confusion. Trajectory responses omit the required
historical trace and evidence updates.}
{fig:case_model_action}

\SVUCaseFigure{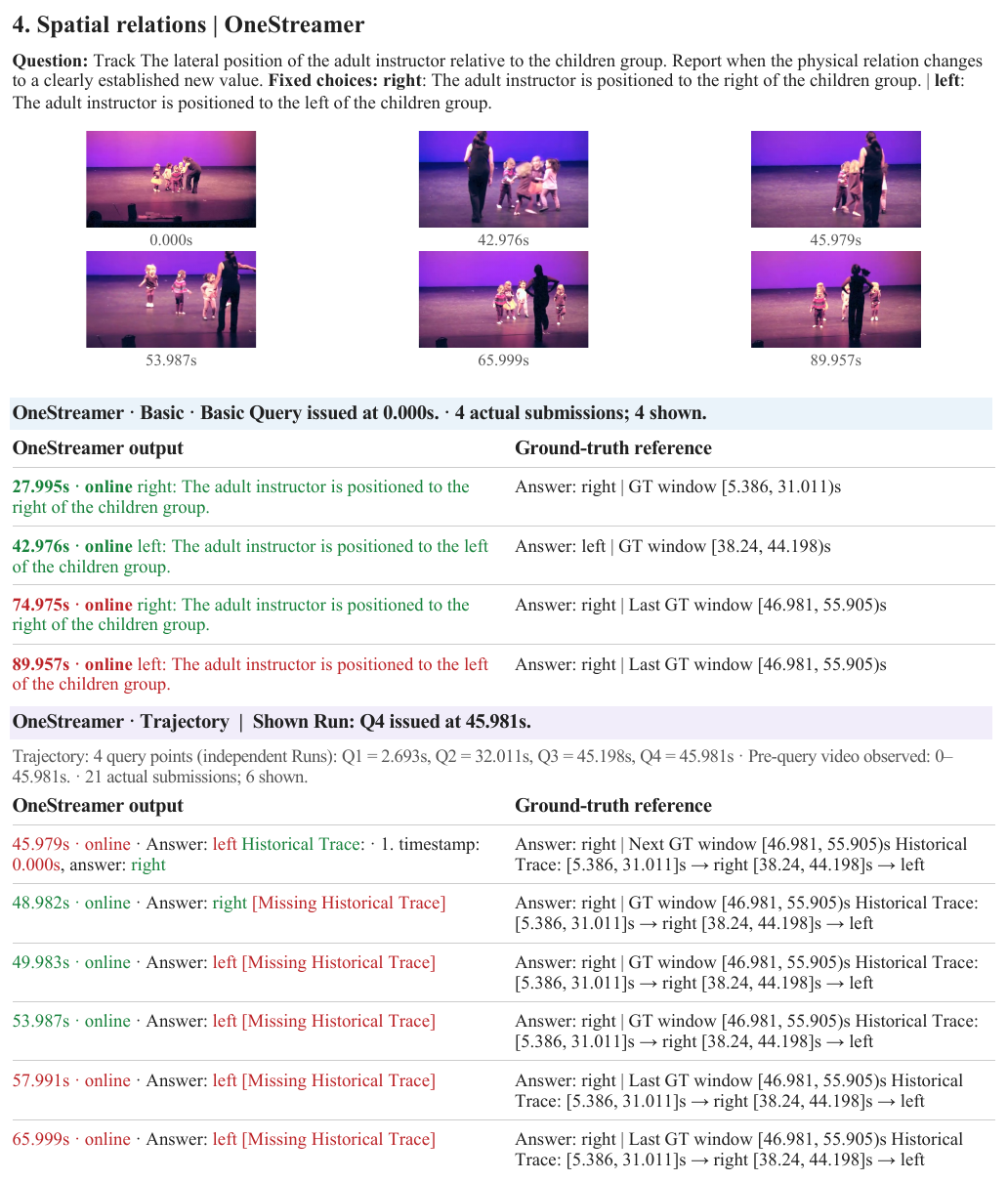}
{\textbf{OneStreamer on Spatial Relation Perception.}
The model produces valid relation reports in Basic mode but
also responds after valid windows. Trajectory mode exhibits
inconsistent answers and incomplete historical reporting.}
{fig:case_model_spatial}

\SVUCaseFigure{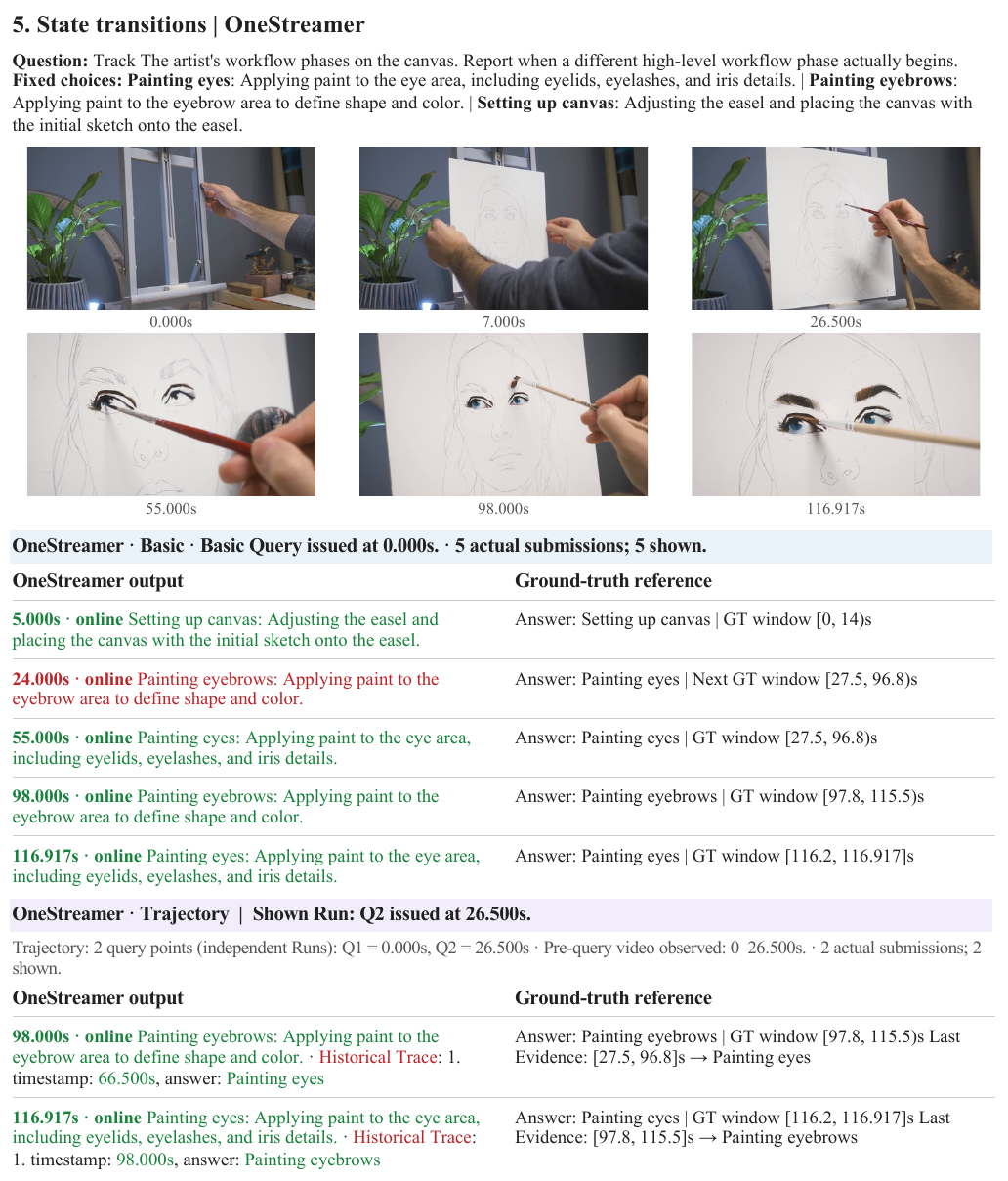}
{\textbf{OneStreamer on State Transition Recognition.}
The model recognizes several workflow phases but also reports
a phase prematurely. Trajectory mode misses an earlier response
window and uses historical fields inconsistently.}
{fig:case_model_state}

\SVUCaseFigure{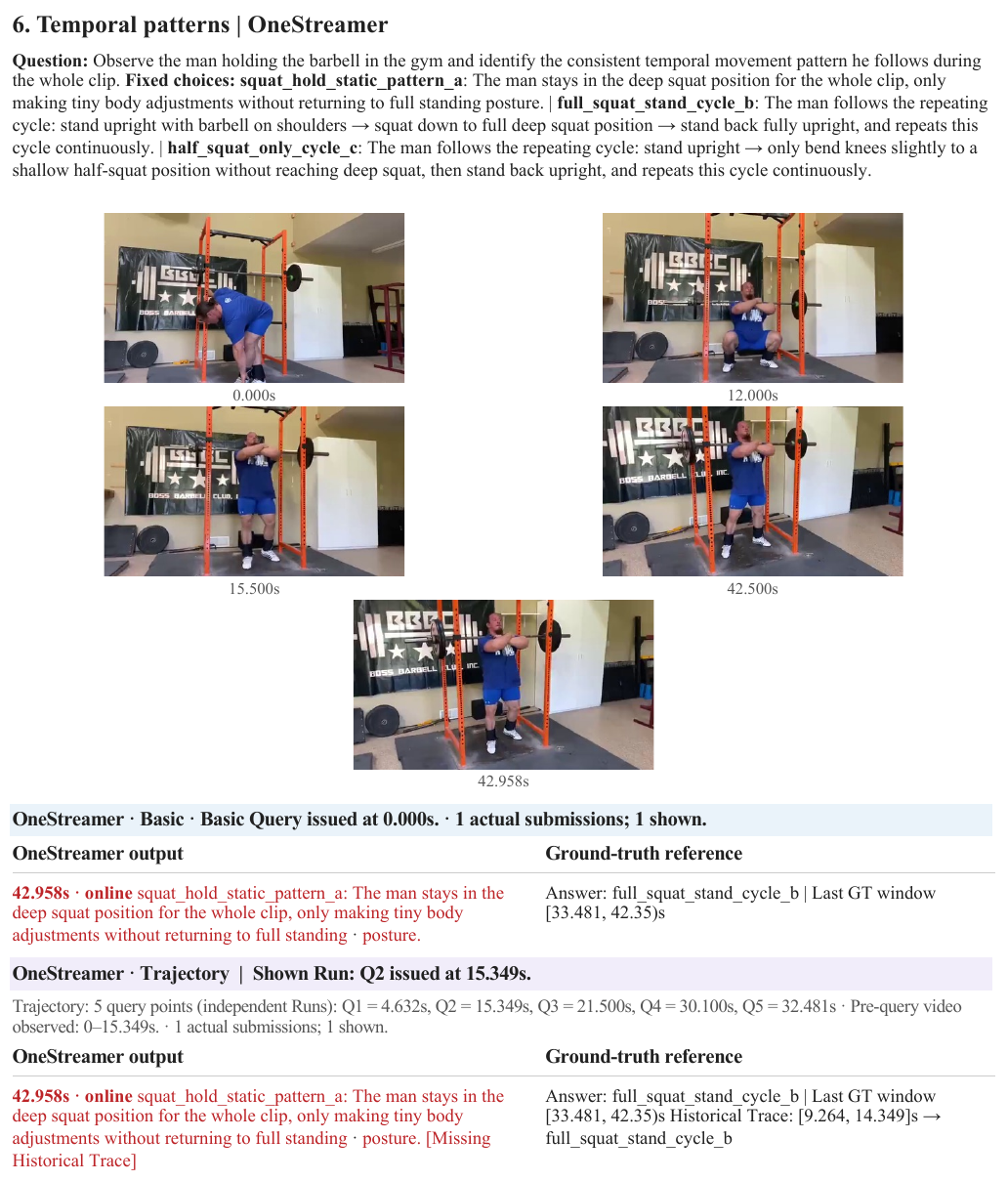}
{\textbf{OneStreamer on Temporal Pattern Recognition.}
The model incorrectly identifies a static posture instead of
the repeated squat cycle and responds after the final valid
window in both evaluation modes.}
{fig:case_model_pattern}

\SVUCaseFigure{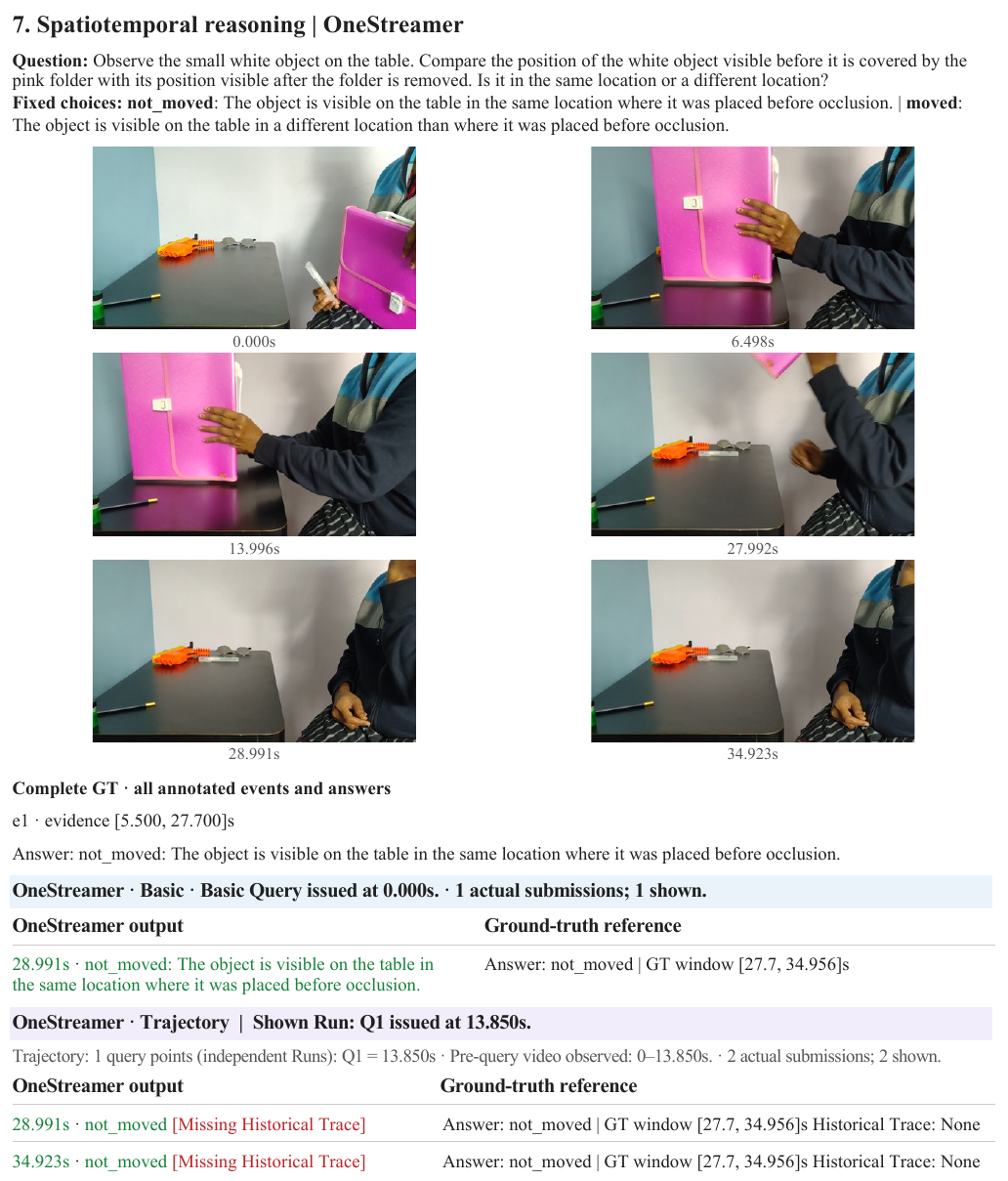}
{\textbf{OneStreamer on Spatiotemporal Reasoning.}
The model correctly compares the object's position before and
after occlusion in Basic mode. Trajectory mode repeats the
answer and omits the required explicit empty historical trace.}
{fig:case_model_spatiotemporal}

\SVUCaseFigure{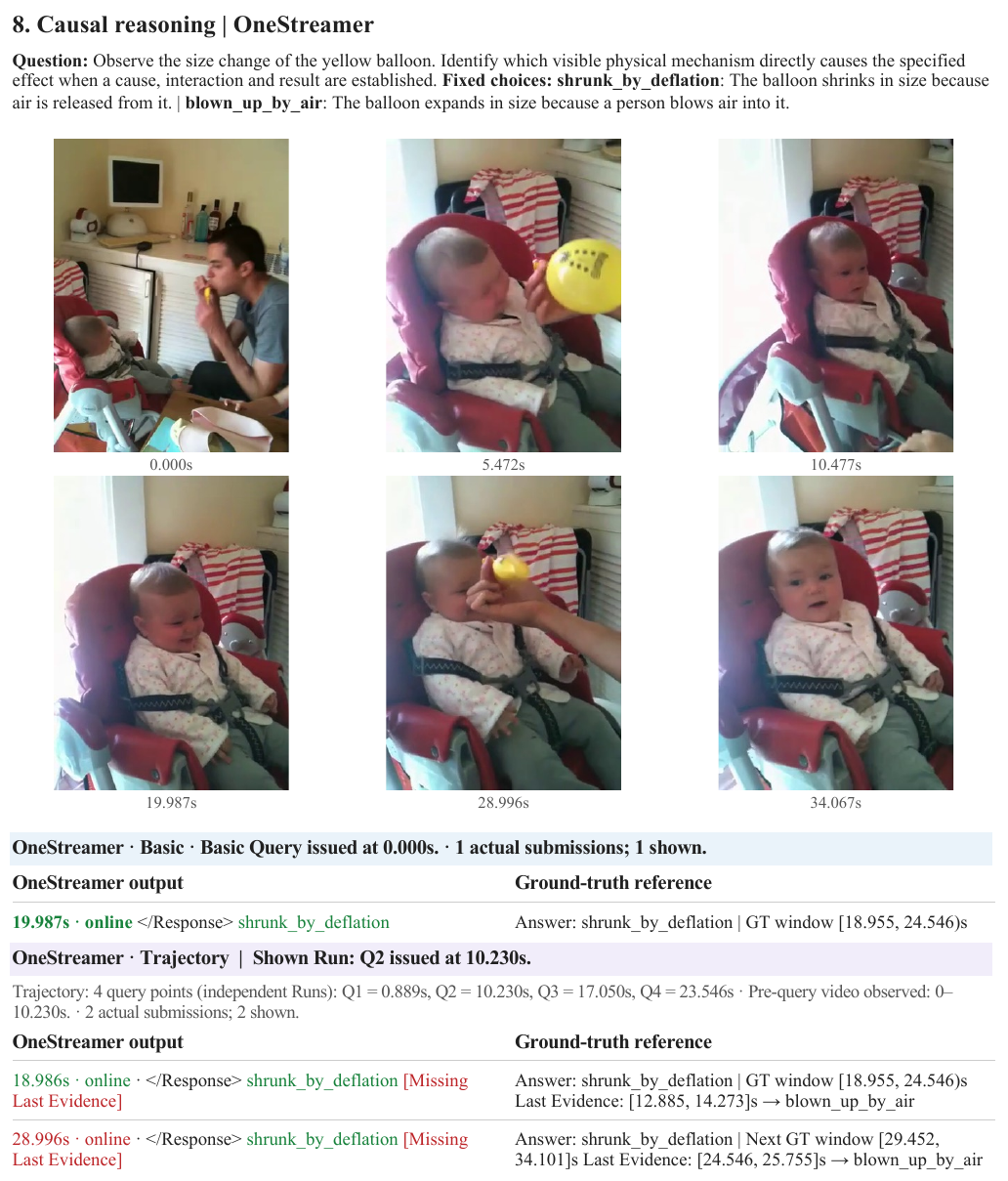}
{\textbf{OneStreamer on Causal Reasoning.}
The model correctly identifies a deflation event but provides
incomplete event coverage. Trajectory responses additionally
exhibit timing errors and missing historical evidence.}
{fig:case_model_causal}

\SVUCaseFigure{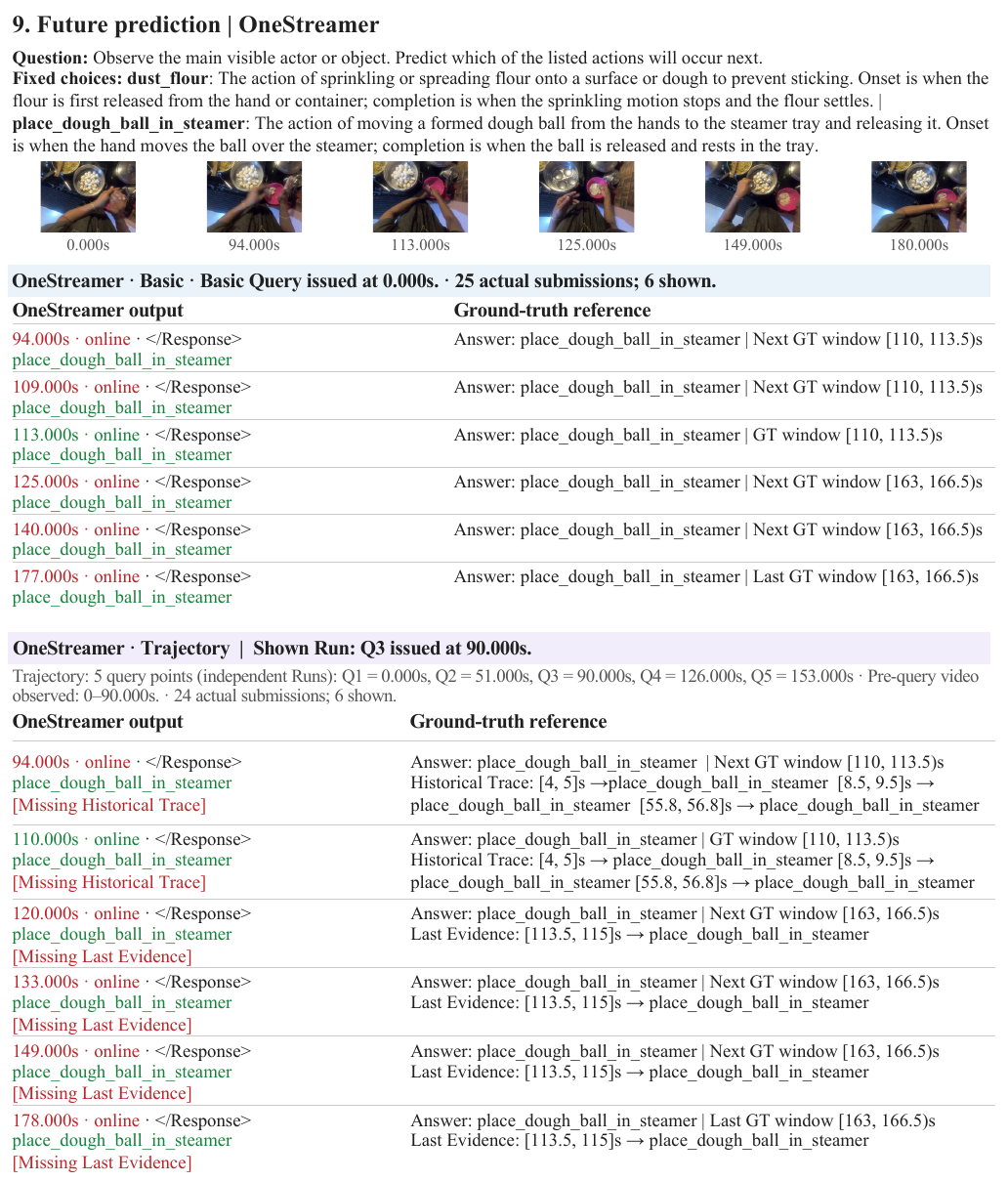}
{\textbf{OneStreamer on Future Prediction.}
Repeated predictions occur outside valid cue windows despite
some correctly timed outputs. Trajectory mode also omits the
required historical trace and evidence fields.}
{fig:case_model_prediction}

\SVUCaseFigure{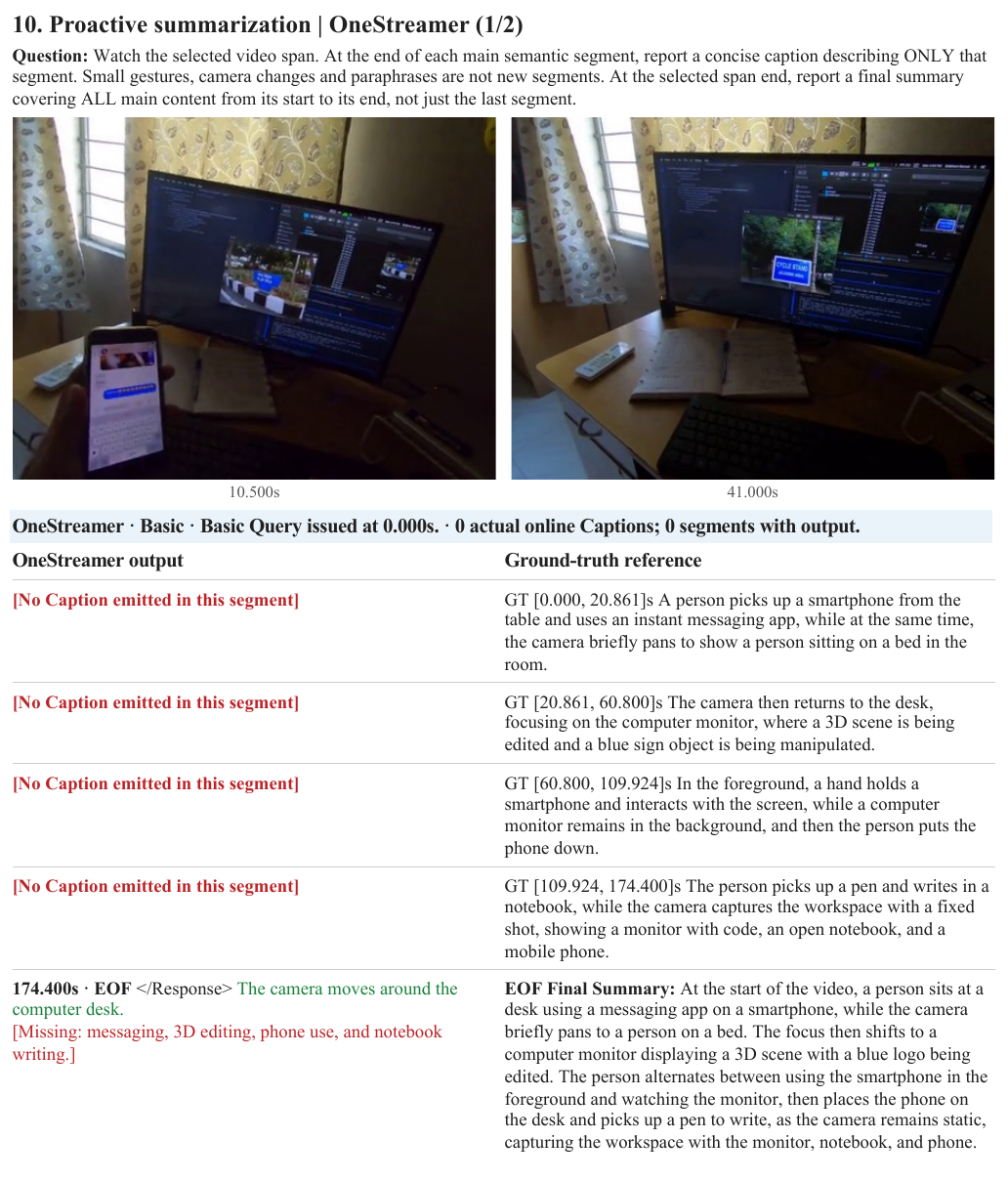}
{\textbf{OneStreamer on Proactive Summarization in Basic mode.}
The model produces no online segment captions, and its final
summary omits the main activities shown in the reference.}
{fig:case_model_summary_basic}

\SVUCaseFigure{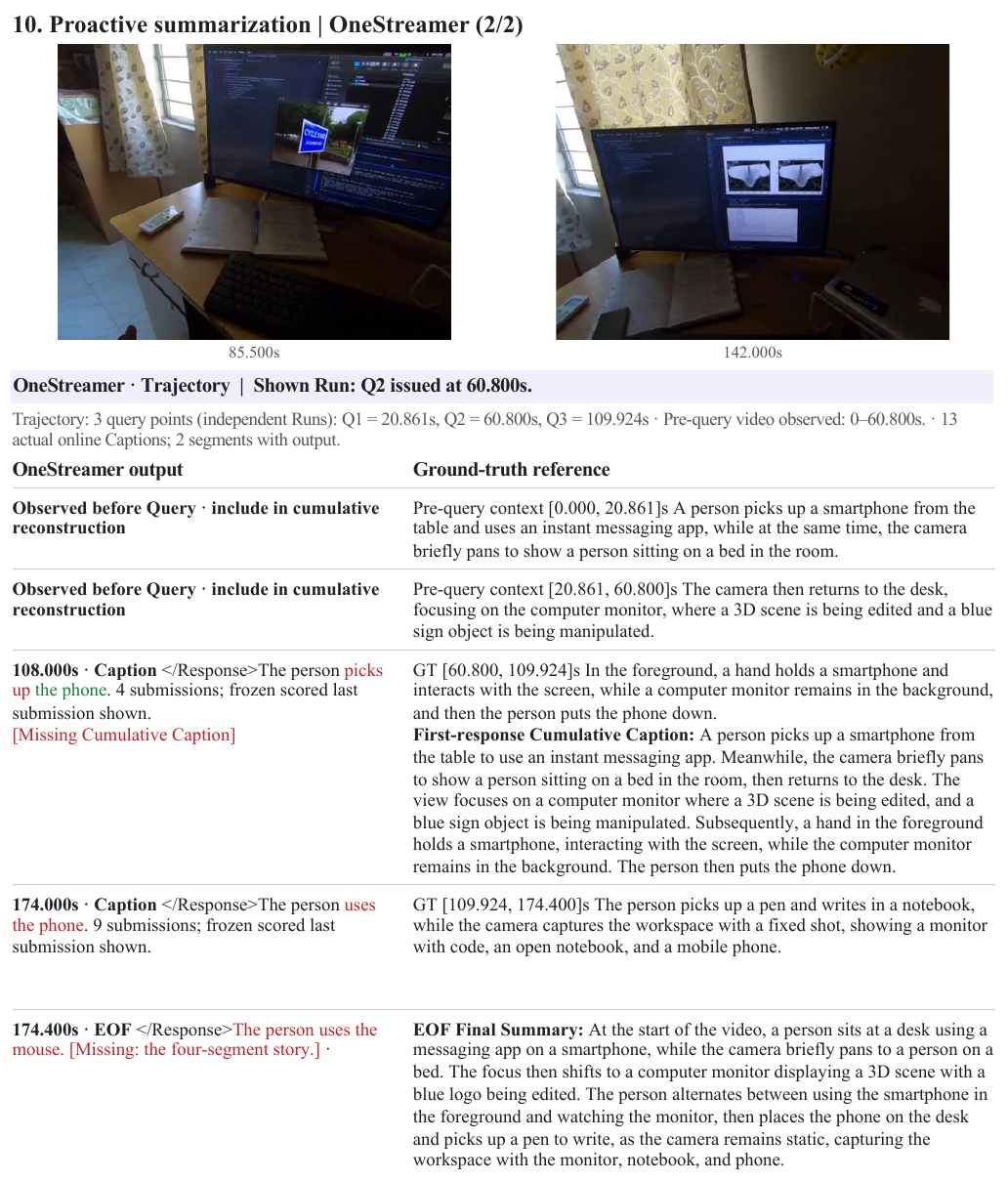}
{\textbf{OneStreamer on Proactive Summarization in Trajectory mode.}
The model produces repeated and partially incorrect captions,
omits the cumulative caption, and fails to summarize the
complete sequence at the end of the video.}
{fig:case_model_summary_trajectory}

\end{document}